\documentclass{article}
\usepackage{arxiv}
\usepackage{graphicx} %
\usepackage{microtype}
\usepackage{subfigure}
\usepackage{booktabs} %
\usepackage{hyperref}

\usepackage{amsmath}
\usepackage{amssymb}
\usepackage{mathtools}
\usepackage{amsthm}
\usepackage{todonotes}
\usepackage{wrapfig}

\usepackage[capitalize,noabbrev]{cleveref}

\newcommand{\eg}{\emph{e.g.}}
\newcommand{\ie}{\emph{i.e.}}

\theoremstyle{plain}

\theoremstyle{definition}

\theoremstyle{remark}

\usepackage{natbib}

\newcommand{\minisection}[1]{\textbf{#1} }

\usepackage[T1]{fontenc}
\usepackage{booktabs}
\usepackage{multirow}
\usepackage{amssymb}
\usepackage{pifont}
\usepackage{lipsum}
\usepackage{tikz}
\usepackage{subfigure}

\usepackage{algorithm}
\usepackage{CJKutf8}

\usepackage{authblk}

\newcommand{\pangu}{Pangu}
\newcommand{\archname}{Pangu Pro MoE}
\newcommand{\modelname}{\archname} %

\title{\archname: Mixture of Grouped Experts for Efficient Sparsity}
\author{\pangu~Team, Huawei\\
pangutech@huawei.com}

\begin{document}
\begin{CJK}{UTF8}{gbsn}

\maketitle
\thispagestyle{fancy}

\begin{abstract}
The surgence of Mixture of Experts (MoE) in Large Language Models (LLMs) promises a small price of execution cost for a much larger model parameter count and learning capacity, because only a small fraction of parameters are activated for each input token. However, it is commonly observed that some experts are activated far more often than others, leading to system inefficiency when running the experts on different devices in parallel. Existing heuristics for balancing the expert workload can alleviate but not eliminate the problem.  Therefore, we introduce Mixture of Grouped Experts (MoGE), which groups the experts during selection and balances the expert workload better than MoE in nature. It constrains tokens to activate an equal number of experts within each predefined expert group. When a model execution is distributed on multiple devices, which is necessary for models with tens of billions of parameters, this architectural design ensures a balanced computational load across devices, significantly enhancing throughput, particularly for the inference phase. Further, we build \modelname~on Ascend NPUs, a sparse model based on MoGE with 72 billion total parameters, 16 billion of which are activated for each token. The configuration of \modelname~is optimized for Ascend 300I Duo and 800I A2 through extensive system simulation studies. Our experiments indicate that MoGE indeed leads to better expert load balancing and more efficient execution for both model training and inference on Ascend NPUs. The inference performance of Pangu Pro MoE achieves 1148 tokens/s per card and can be further improved to 1528 tokens/s per card by speculative decoding on Ascend 800I A2, outperforming comparable 32B and 72B Dense models. Furthermore, we achieve an excellent cost-to-performance ratio for model inference on Ascend 300I Duo. Our studies show that Ascend NPUs are capable of training \modelname~with massive parallelization to make it a leading model within the sub-100B total parameter class, outperforming prominent open-source models like GLM-Z1-32B and Qwen3-32B. \footnote{\url{https://gitcode.com/ascend-tribe/pangu-pro-moe}
.}
\end{abstract}

\section{Introduction}
\label{sec:intro}
The Mixture-of-Experts (MoE) model~\cite{shazeer2017outrageously, cai2024survey, jiang2024mixtral} is becoming a standard component in  Large Language Models (LLMs)~\cite{du2022glam,touvron2023llama,wang2023pangu,liu2024deepseek,Yin2025PanguUP}, where scaling model size brings clear benefits. MoE models only activate a subset of experts for each token to effectively reduce the computation cost of the gigantic language models. However, expert load imbalance~\cite{lepikhin2021gshard,yang2024qwen2} is a critical challenge for materializing the computation gain on distributed training and inference systems. Because of the gigantic scale of recent LLMs, the model parameters are usually loaded separately on different computing devices, like Ascend NPUs~\cite{tang2025panguultramoetrain}. It is commonly observed that some experts are frequently activated for input tokens while some remain rarely used. Since experts are typically distributed across multiple devices, the device with busiest experts straggle both training and inference processes, and further hinder computational efficiency and throughput.

To overcome this limitation, we develop the Mixture of Grouped Experts (MoGE) architecture. When selecting the experts for a token, we divide the experts into equal groups and then choose experts from each of the groups. Each group has an identical number of activated experts. In typical distributed deployments, the experts are assigned to the devices according to the group ID. MoGE effectively balances the computational load across all participating devices. This design offers substantial improvements in throughput in training and inference scenarios.

Based on MoGE, we build \modelname, with 72 billion parameters, 16 billion of which are activated for each input token. We configure the model structure based on extensive simulation studies, aiming for optimized runtime performance on Ascend 300I Duo and 800I A2 platforms. The inference strategy is also optimized for \modelname~on multiple aspects, including system, algorithms, and kernel design, specifically tailored for the Ascend NPUs. At the system level, we propose a hierarchical \& hybrid parallelism and communication strategy, co-designed with the model architecture and Ascend’s interconnect topology. This design significantly reduces redundant computation and communication overhead. Compression algorithms such as quantization are also applied to further reduce computational cost. At the kernel level, we develop high-performance MulAttention and SwiftGMM kernels, custom-optimized for Ascend NPUs. The experiments demonstrate that the optimized inference of \modelname, configured as 72BA16B MoE, on the Ascend NPUs, achieves low latency in low-concurrency scenarios and high throughput in high-concurrency settings, outperforming comparable dense models like 32B and 72B dense models.

\modelname~is pre-trained on a diverse and high-quality corpus comprising 13 trillion tokens using 4K Ascend NPUs. Following this extensive pre-training phase, we employ Supervised Fine-Tuning (SFT) and subsequent Reinforcement Learning (RL) to further enhance its reasoning capabilities. Our comprehensive evaluation results shows \modelname~is a forefront model with total parameters under 100 billion. It outperforms several strong open-source models, including GLM-Z1-32B~\cite{glm2024chatglm}, Qwen3-32B~\cite{yang2025qwen3}, and Gemma3-27B~\cite{team2025gemma}, across a wide range of competitive benchmarks.

\begin{figure}[t]
    \begin{center}
\centerline{\includegraphics[width=0.9\columnwidth]{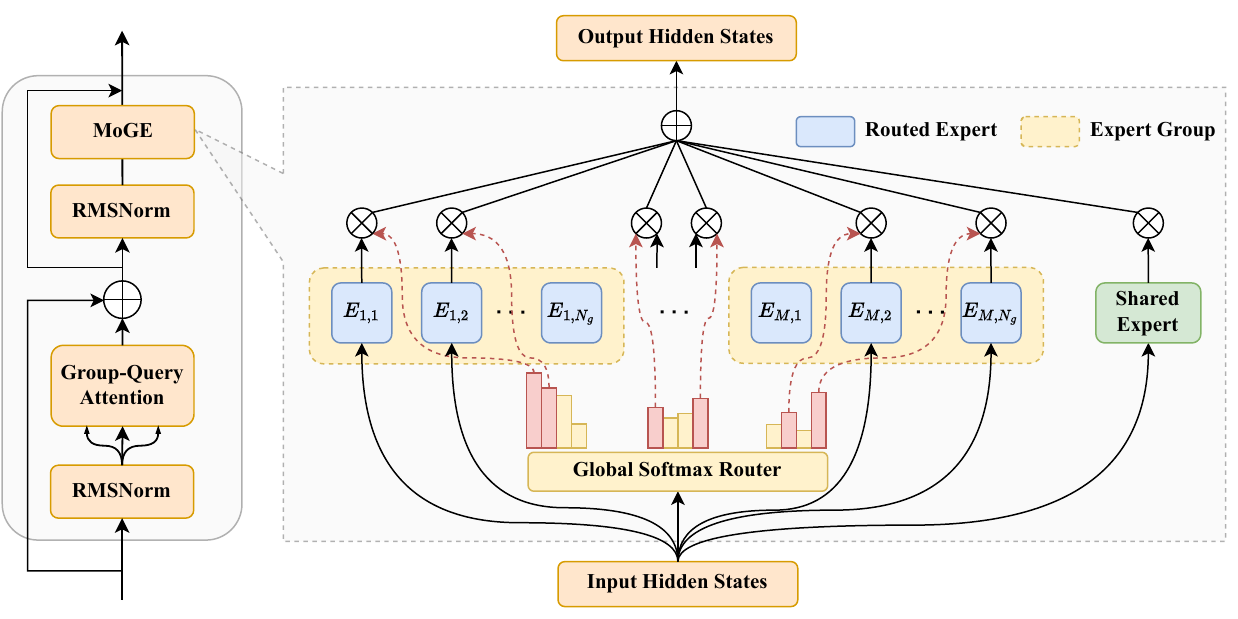}}
    \caption{Illustration of the Mixture of Grouped Experts (MoGE) architecture. The $N$ total experts are evenly partitioned into $M$ non-overlapping groups, with each group typically assigned to a distinct computational device. For each input token, initial gating scores for all experts are computed via a global softmax router. Subsequently, within each expert group, the $K^{\prime}$ experts with the highest scores are chosen based on these initial scores. The scores corresponding to unselected experts are effectively set to zero. The final output is obtained by a weighted sum of the outputs from the activated experts and a shared expert.}
    \label{img: arch}
    \end{center}
\end{figure}

\section{Mixture of Grouped Experts}
\label{method}

In this section, we introduce the MoGE, a novel MoE architecture designed to intrinsically achieve perfect load balancing across devices. MoGE achieves this by partitioning experts into distinct groups, with each group typically mapped to a specific device, and then enforcing a routing strategy that activates a fixed number of experts from each group. We begin by revisiting the expert imbalance problem in conventional MoE, then detail the MoGE architecture and its associated routing mechanisms, and finally discuss the auxiliary losses designed to optimize its performance.

\subsection{Expert Imbalance in Conventional MoE}

We consider an input hidden state $\boldsymbol{h}$ from the space $\mathcal{X} \subseteq \mathbb{R}^d$. The MoE layer comprises $N$ distinct expert networks, denoted as $\{E_1, E_2,\cdots, E_N\}$ where each $E_i: \mathcal{X} \rightarrow \mathcal{Y}$ maps input hidden states to output representations in space $\mathcal{Y}$. The output of the MoE module for a given input hidden states $\boldsymbol{h}\in \mathcal{X}$ is the weighted sum over the $N$ expert networks:
\begin{equation}
    \boldsymbol{y} = \sum_{i = 1}^N G(\boldsymbol{h})_i \cdot E_i(\boldsymbol{h}),
    \label{eq:moe_output}
\end{equation}
where $G(\boldsymbol{x})_i$ represents the gating score, or weight, assigned by a router mechanism to the $i$-th expert $E_i$. These scores determine how much influence each expert has on the final output. Typically, $G(\boldsymbol{h})$ is implemented by taking the softmax over the Top-K~\cite{shazeer2017outrageously} inner-product between the router matrix $\boldsymbol{W} \in \mathbb{R}^{d \times n}$ and $\boldsymbol{h}$:
\begin{equation}
\begin{aligned}
G(\boldsymbol{h}) &  = \mathtt{Softmax} \left(\mathtt{TopK}\left(\boldsymbol{W}^{\top}\boldsymbol{h}, K \right )\right), \\
\mathtt{TopK}(\boldsymbol{v}, k) &  = \begin{cases} 
v_i,  & \text{if $v_i$ is in the top $k$ values of $\boldsymbol{v}$.}  \\
-\infty, & \text{otherwise.}
\end{cases},
\\
\end{aligned}
\label{eq:topk_routing}
\end{equation}
where $\mathtt{Softmax}(\boldsymbol{u}) = \exp(u_i) \big / \sum_i \exp(u_i) $ and $\mathtt{TopK}(\boldsymbol{v}, k)$ selects the top $k$ elements based on the values in $\boldsymbol{v}$. $K$ denotes the number of experts activated per token. This routing method allows for sparse activation of experts by skipping computations for expert $E_i$ where $G(\boldsymbol{h})_i = 0$.

\begin{figure}[t]
    \centering
	\subfigure[The distribution of activated experts for a single token.]{
	    \includegraphics[width=0.6\linewidth]{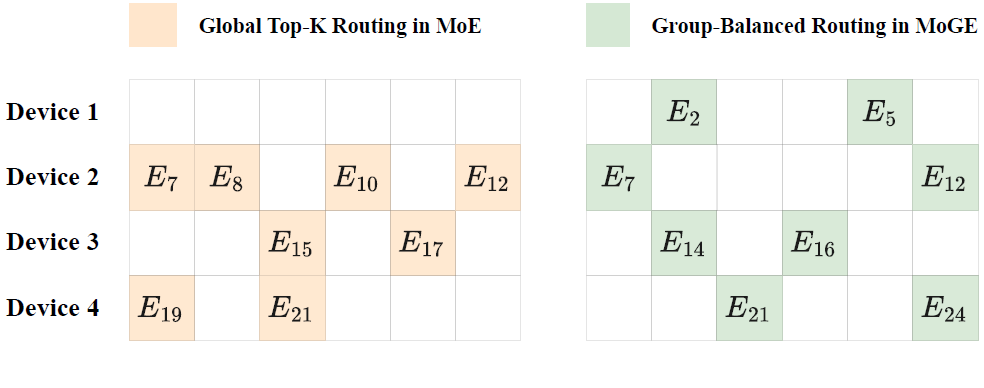}
	    \label{img: balanced_vs_topk}}
	\vspace{0.4cm}\subfigure[Distribution of imbalance score.]{
	\includegraphics[width=0.3\linewidth]{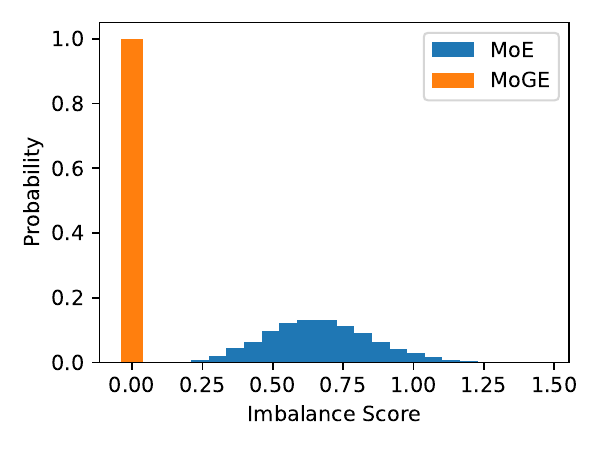}
        \label{img: modeling_imbalance}
	}
	\caption{Comparison of expert activation patterns and load imbalance between conventional Top-K routing in MoE and MoGE's group-balanced routing. \textbf{(a)} Illustrates how experts are selected for a single input token. In this example, 8 experts are chosen from a total of 24 experts, with an expert parallelism of 4. Conventional Top-K routing (left) selects 8 experts that are unevenly distributed across the 4 devices (\ie, Device 1 gets zero experts while Device 2 gets 4). In contrast, MoGE (right) divides the 24 experts into 4 groups and selects exactly 2 experts from each group. \textbf{(b)} Shows the estimated probability distribution of the Imbalance Score (IS) for both routing mechanisms, where a lower IS indicates better balance. MoGE inherently achieves an IS of 0, signifying perfect load balance. Conventional Top-K routing, however, exhibits a high probability of IS values greater than 0, indicating frequent load imbalance.}
\label{img: advantages}
\end{figure}

In practical implementations, especially for large models with tens of billions of parameters, these $N$ experts are typically distributed evenly across $M$ computational devices (\eg, NPUs) to enable parallel processing. However, the standard Top-K routing mechanism described in Eq.~\eqref{eq:topk_routing} places no constraints on which $K$ experts are selected. For a given token, the $K$ chosen experts could, by chance, be concentrated on a few devices, as depicted in Figure~\ref{img: balanced_vs_topk} (left).

This uneven assignment of active experts to devices leads to a load imbalance. Some devices might be busy processing many tokens (if their resident experts are frequently chosen), while others might be idle or underutilized. This creates a "straggler" problem: the overall processing speed is dictated by the slowest (most heavily loaded) device, leading to inefficient use of computational resources and increased latency.

To quantify this load imbalance, we introduce the Imbalance Score (IS). Imagine we are processing a batch of input data (\eg, a set of tokens) $X$. For this batch, let $T_i(X)$ be the total number of expert computations (or tokens routed to experts) handled by device $i$. The Imbalance Score is then defined as the difference between the maximum and minimum load across all $M$ devices, normalized by the batch size $|X|$:
\begin{equation}
   IS(X)  = \frac{1}{|X|}{\left [\max_{i \in \{1, \cdots, M\}}T_i(X)-\min_{i \in \{1, \cdots, M\}}T_i(X)\right ]}.
   \label{eq:imbalance_score}
\end{equation}
A higher IS value signifies a greater disparity in workload among devices, indicating a more severe imbalance. Conversely, an IS of 0 implies perfect load balance, where every device processes an equal amount of work. This metric is crucial because it directly reflects the potential for performance bottlenecks due to uneven workload distribution. Even if the average load per device is reasonable, a high IS means some devices are overloaded while others are waiting, reducing overall system efficiency.

To understand the propensity of Top-K routing towards imbalance, we can estimate the distribution of the Imbalance Score. If we assume, for simplicity, that each of the $K$ experts chosen for a token is selected independently and uniformly at random from the $N$ total experts, we can use Monte Carlo simulation. For a given set of parameters, we can simulate the routing process many times (say, $D$ times) and calculate the IS for each simulation. The probability of observing a specific IS value $v$ can then be estimated as:
\begin{equation}
  P(IS=v) = \frac{1}{D} \sum_{j = 1}^{D} \mathbb{I}\left(IS(X_j) = v\right),
\end{equation}
where $X_j$ is the $j$-th simulated batch, and $\mathbb{I}(\cdot)$ denotes the indicator function. Figure~\ref{img: modeling_imbalance} (right, blue bars) illustrates such an estimated distribution for a configuration with $N=64$, $K=8$, $M=8$, and a small batch size of $|X|=16$ tokens. The distribution clearly shows that non-zero IS values are highly probable. In fact, for these parameters, the probability of $IS >0$  (\ie, some level of imbalance) is nearly 1. This signifies that with standard Top-K routing, load imbalance is almost an inevitable occurrence, especially when the batch size is small, as there's less opportunity for random selections to average out perfectly across devices.

\subsection{Basic Architecture of MoGE}
\label{MoGE}

To directly tackle the load imbalance problem inherent in conventional MoE, we propose the Mixture of Grouped Experts (MoGE) architecture, which employs a novel group-balanced routing strategy. The core idea is to ensure that for each token, an equal number of expert computations are distributed to each of the devices. This guarantees an Imbalance Score (IS) of 0 by design, as illustrated by the single yellow bar in Figure~\ref{img: modeling_imbalance}.

The MoGE architecture (visualized in Figure~\ref{img: arch}) achieves this as follows:
\begin{enumerate}
    \item \textbf{Expert Partitioning}: The $N$ total experts are deterministically partitioned into $M$ distinct, non-overlapping groups. Each group, say group $j$ (where $j = 1,2,\cdots,M$), contains $N_g = N / M$ experts. Crucially, each group of experts is typically assigned to reside on a specific computational device.  
    \item \textbf{Group-Balanced Routing}: For each input token $\boldsymbol{h}$, the routing mechanism ensures that a fixed number of $K^{\prime} = K / M$ experts are activated from each of the $M$ groups. This means a total of $K$ experts are also activated per token, but now with a strict per-device activation count.
\end{enumerate}

First, similar to standard MoE, we compute initial scores for all $N$ experts using the input token $\boldsymbol{h}$ and a router weight matrix $\boldsymbol{W} \in \mathbb{R}^{d \times n}$. Instead of directly applying a global Top-K, we first apply a \emph{global softmax} to these raw scores to obtain normalized probabilities or affinities, $\boldsymbol{S}$, for all experts:
\begin{equation}
    \boldsymbol{S} = \mathtt{Softmax}\left(  \boldsymbol{W}^{\top}\boldsymbol{h} \right)
    \label{global gating score}
\end{equation}
Here, $\boldsymbol{S}$ is a vector of length $N$, where $S_i$ is the initial score for expert $E_i$. 

Next, we consider these scores group by group. Let $\boldsymbol{S}_j$ denote the sub-vector of $\boldsymbol{S}$ containing the scores for the $N_g = N / M$ experts belonging to group $j$. Within each group $j$, we perform a \emph{local Top-$K^{\prime}$} selection based on the scores in $\boldsymbol{S}_j$.  The gating scores $G^{\prime}(\boldsymbol{h})$ for MoGE are then constructed by concatenating the results of these local Top-$K^{\prime}$ operations from all $M$ groups:
\begin{equation}
G^{\prime}(\boldsymbol{h})  =  \left (\mathtt{TopK}\left(\boldsymbol{S}_1, K^{\prime}\right),  \cdots, \mathtt{TopK}\left(\boldsymbol{S}_M, K^{\prime}\right) \right ).
\label{eq: vanilla grouping}
\end{equation}
Effectively, for each group $j$, we select the $K^{\prime}$ experts with the highest scores within that group. Experts not selected within their group receive a weight of 0. A common and particularly impactful configuration is when we want to select exactly one expert from each group, meaning $K^{\prime} = 1$. In this case, Eq.~\eqref{eq: vanilla grouping} simplifies to performing a $K$ Top-1 selection within each group. \modelname~follows this setting as shown in Table~\ref{tab:model_config}. 

The final output $\boldsymbol{y}_{\text{MoGE}}$  is then computed similarly to Eq.~\eqref{eq:moe_output}, but using these group-derived gating scores:
\begin{equation}
\boldsymbol{y}_{\text{MoGE}} = \sum_{i=1}^{N} G^{\prime}(\boldsymbol{h})_i \cdot E_i(\boldsymbol{h}).
\label{eq:moge_output}
\end{equation}

\noindent
\minisection{Auxiliary Load Balancing Loss}
While MoGE structurally guarantees inter-group load balance (\ie, across devices), it is still crucial to ensure that the routing mechanism learns to distribute the workload reasonably among experts within each group. To achieve this, we employ a batch-level auxiliary load balancing loss:
    \begin{equation*}
    \ell_{\text{aux}} = \alpha \sum_{i = 1}^N f_i p_i,
    \end{equation*}
where the hyperparameter $\alpha$ controls the strength of the auxiliary loss. Here, $f_i$ represents the fraction of tokens within the batch $\mathcal{B}$ routed to expert $E_i$, and $p_i$ is the average gating score assigned to expert $E_i$:
\begin{equation}
f_i  = \frac{N}{K |\mathcal{B}|} \sum_{t \in \mathcal{B}} \mathbb{I} \left \{ \text{Token $t$ selects Expert $i$}\right \}, \\ \quad 
p_i  = \frac{1}{\mathcal{B}} \sum_{t \in \mathcal{B}} S_{i, t},
\label{selection}
\end{equation}
where $\mathbb{I}\{\cdot\}$ is the indicator function, and $S_{i, t}$ denotes the gating score of expert $E_i$ for token $t$ (derived from the global softmax scores $\boldsymbol{S}$ in Eq.~\eqref{global gating score} before group-wise Top-K selection). Our ablation studies indicate that computing this auxiliary balancing loss based on the global softmax weights (\ie, considering all experts collectively for the loss signal) yields superior performance compared to calculating the loss independently and locally for each group based on its intra-group routing decisions.

\begin{figure}[t]
    \centering
    \includegraphics[width=0.6\linewidth]{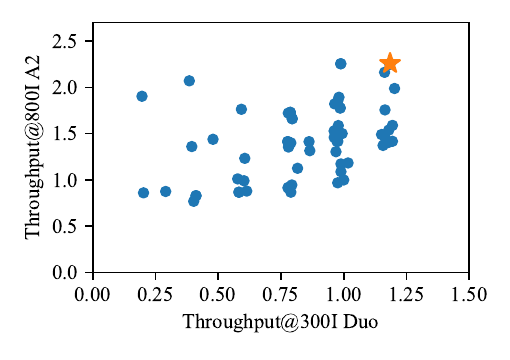}
    \vspace{-0.5cm}
\caption{Simulation results of candidate model configurations. Throughput under 100 ms and 50 ms TPOT constraints on Ascend 300I Duo and 800I A2 platforms, respectively, is recorded. All results are normalized relative to a randomly selected candidate.}
\label{img: modeling_config}
\end{figure}

\subsection{Architecture Simulation for Ascend NPUs}

This section primarily focuses on how we determine the key hyperparameters of the model to ensure better affinity with Ascend hardware. During the model design process, a hierarchical strategy is employed, progressing from coarse to fine granularity to balance accuracy and inference efficiency on Ascend 300I Duo and 800I A2 platforms. The strategy comprises three stages: first, a coarse-grained filter determines the parameter range based on memory bandwidth and latency constraints of a single server; second, domain expertise is used to shortlist potential models, narrowing the design space; third, the performance of candidate models is evaluated using an operator-level simulator that correlates system hardware parameters, such as TFLOPS, memory access bandwidth, memory capacity, and interconnection topology, and automatically searches for optimal parallelism.

\begin{wraptable}{r}{0.5\linewidth}
\vspace{-0.2cm}
\centering
\caption{Model configuration for \modelname.}
\label{tab:model_config}
    \begin{tabular}{lc}
    \toprule
    Configuration & \modelname \\
    \midrule
    Vocabulary Size & 153376 \\
    Hidden Size  & 5120 \\
    Intermediate Size    & 1344 \\
    Query Heads  & 40 \\
    KV Heads   & 8 \\
    Head Size    & 128 \\
    Layers    & 48 \\
    Routed Experts   & 64 \\
    Activated Experts     & 8 \\
    Shared Experts    & 4 \\
    \midrule
    \# Activated Parameters (B) & \textbf{16.50} \\
    \# Total Parameters (B) & \textbf{71.99} \\
    \bottomrule
    \end{tabular}%
    \vspace{-0.2cm}
\end{wraptable}

Figure~\ref{img: modeling_config} presents the simulation results for candidate models. These models are configured within specific parameter ranges, including hidden dimensions (4096-8192), query heads (32-64), key-value heads (8-16), number of layers (40-64) and routed expert counts (32–64). All configurations maintain a total parameter count within the range of 60–80 billion.

During the decoding phase, the inference performance is primarily influenced by memory access and communication overhead. The model architecture, particularly the width-to-depth ratio, significantly impacts communication efficiency. For example, the hidden size determines the data volume per communication, while the number of layers affects the communication frequency. The trade-off between these two factors can be effectively evaluated by considering the system's static communication latency and available bandwidth. From the memory access perspective, under identical parameter scales, a higher sparsity ratio reduces the number of parameters that need to be loaded under the same concurrency conditions. Through the hierarchical strategy and fine-grained simulation, the orange star-marked model in Figure~\ref{img: modeling_config} exhibits superior performance for Ascend 300I Duo and 800I A2 platforms in all candidates under the specified conditions, which serves as the model configuration for Pangu Pro MoE, as shown in Table~\ref{tab:model_config}.

\section{Model Training}
\label{train}
In this section, we present the training pipeline for \modelname. Our framework comprises two key phases: pre-training and post-training. Each phase employs specific training strategies and data curricula to progressively enhance the model capabilities. The model configuration of \modelname~is shown in Table~\ref{tab:model_config}.

\subsection{Pre-Training}

\label{pretrain}
We first detail the data construction used in the pre-training phase, emphasizing their relevance to model capabilities. Next, we explain the pre-training strategies that ensure stable model convergence.

\subsubsection{Data Construction} 
The pre-training dataset of \modelname~contains a total number of 13 trillion tokens produced by our tokenizer with a vocabulary size of 153,376 tokens. The tokenizer is designed by a domain-aware vocabulary strategy that aims to maintain balanced representation between domains, and more details can be found in~\cite{Yin2025PanguUP}. This dataset contains diverse and high-quality content from various sources, such as web pages, books, multilingual, code, STEM (Science, Technology, Engineering, and Mathematics), industrial domains, reasoning and synthetic data.

\paragraph{Training Phases} The overall pre-training process of \modelname~is structured into three sequential phases grounded in cognitive development theories: the \textit{general} phase (9.6T), the \textit{reasoning} phase (3T), and the \textit{annealing} phase (0.4T). The three phases are designed to progressively develop core capabilities of \modelname, such as general knowledge and linguistic ability in the first phase, then to improve reasoning skills of the model in the second phase, and to further refine model knowledge and behavior in the third phase. 
Apart from the general data from various sources, we particularly involve a lot of high-quality data from multiple industrial domains in the first \textit{general} phase.
In general, the first phase is trained with a 4K sequence length, and the latter two are trained with a 32K sequence length.

The second \textit{reasoning} phase targets the reasoning skills of \modelname~by significantly increasing the proportion of more complicated data such as STEM, coding and internal data. 
We put great effort into the amount and quality of the reasoning data, by optimizing the data cleaning, data generating, and data evaluating pipeline. 
We particularly design synthetic short and long chain-of-thought (CoT) for those difficult samples. To better align with long CoT responses, we use a 32K sequence length. In addition, \modelname~is trained with a significantly larger range of high-quality reasoning data than the previous model~\cite{tang2025panguultramoetrain, Yin2025PanguUP}, which therefore helps it achieve good performance even in a relatively small model size.

The third \textit{annealing} phase serves to bridge the transition from pre-training to post-training, where the instruction style data increases to approximately 20\% of the corpus. In this phase, priority is given to the data with extremely higher quality and difficulty scores, following a curriculum-based sampling strategy throughout all three phases. We also intentionally increase the number of data on an advanced level of STEM education, which is totally 18\% of the corpus. Using a proxy model with 7 billion parameters, we perform intensive ablation studies on data selection and data recipe in this phase, aiming to evaluate how different strategies affect the model. Our training shows that, by equipping itself with a proper data recipe and decayed learning rate, this third phase can still bring about a large improvement in model performance. 

\paragraph{Data Evaluation} The quality of our training corpus is continuously monitored and improved using our domain-aware model-based evaluation, where we fine-tune some Pangu series models as evaluators. Specifically, we design annotated datasets in this fine-tuning process for different domains and thus create a few domain-aware evaluators. Our ablation study on the small proxy model shows that this data evaluation system produces better evaluation performance than using only one unified evaluator. All of our data samples are sent through this evaluation system and assigned scores in multiple dimensions, including cleanliness, fluency, educational value, and richness. These scores are used in our data selection and sampling strategy.

In addition, a category label is given to each data sample, with 188 categories in total. This data label acts as a classifier, making it easier to classify and group data, and more importantly, it optimizes the data mixture in a fine-grained manner and ensures that our training corpus covers a reasonable distribution of various topics.

\subsubsection{Pre-training Parameters}

\paragraph{Training Details} During the pre-training stage, our models are trained for a single epoch using the AdamW optimizer with hyperparameters $\beta_1 = 0.9$ and $\beta_2 = 0.95$. A cosine learning rate schedule is employed throughout, encompassing three progressive phases. In the \textit{general} phase, the learning rate decays from $3 \times 10^{-4}$ to $3 \times 10^{-5}$ with a batch size of 4 million tokens. This is followed by the \textit{reasoning} phase, during which the learning rate further decreases from $3 \times 10^{-5}$ to $1 \times 10^{-5}$, and the batch size is increased to 16 million tokens to enhance training on complex reasoning tasks. In the final \textit{annealing} phase, the learning rate is gradually reduced from $1 \times 10^{-5}$ to $1 \times 10^{-7}$, with the batch size maintained at 16 million tokens to ensure stable convergence. This structured, multi-phase training strategy facilitates both robust generalization and effective specialization across different learning objectives.

\paragraph{Training Devices} The proposed architecture is trained and evaluated using the Huawei Ascend 800T A2. The Ascend 800T A2 is a high-efficiency AI server designed with Huawei's proprietary DaVinci architecture~\cite{liao2019davinci}. A single accelerator in Ascend 800T achieves a computational throughput of 256 TeraFLOPS for half-precision floating-point (FP16) operations and 512 TeraOPS for integer precision (INT8) calculations. Despite its superior performance, it has a maximum power consumption of only 310W, significantly lower than its design specification of 350W. The integration of diverse computing units within the DaVinci architecture enhances the completeness and efficiency of AI computations, thereby extending its applicability and significantly improving the overall performance of AI systems while reducing deployment costs.

\subsection{Post-Training Alignment} %
\minisection{Supervised Fine-Tuning}
The post-training supervised finetuning data for \modelname~is categorized into reasoning and non-reasoning subsets, with a sampling ratio of 3:1 in favor of reasoning tasks. Specifically, the reasoning samples primarily includes tasks such as mathematical problem-solving, code generation, and logical inference, while the non-reasoning samples focuses on general language instruction following, question answering, text generation, long-context understanding, semantic classification, and tool usage.
The reasoning-intensive tasks, especially those involving multi-step computation, symbolic manipulation, or formal logic, are often underrepresented in instruction-tuning corpora, yet they are critical for the emergence of slow thinking capabilities in large language models. By allocating more weight to such tasks, we aim to explicitly encourage the model to develop robust intermediate reasoning skills and deeper cognitive representations that generalize beyond surface-level pattern matching.
For each sub-domain, we utilize a simple yet effective diversity-based metric~\cite{yin2024entropy} to select representative instructions from a large pool of high-quality, expert-curated, and synthetic instruction data. This diverse task composition constitutes a multi-dimensional training space, designed to enhance the model’s generalization capability across both specialized and general-purpose tasks.

Furthermore, the training adopts a two-stage progressive optimization strategy over six training rounds, with global batch sizes of 64 and 32 in each respective stage. The phased design allows for a curriculum-like learning process, where the model initially focuses on broader instruction-following behavior and gradually transitions to mastering more complex reasoning tasks with finer gradient updates. This progressive training schedule not only stabilizes convergence but also reduces catastrophic forgetting when reasoning tasks are emphasized in later rounds.
From the perspective of a slow thinking model, this design provides two key advantages in order for the model to have stronger reasoning capabilities.
First, emphasizing reasoning early on builds strong inductive biases toward stepwise problem solving. Second, the staged finetuning allows the model to consolidate general linguistic capabilities before being pushed toward more cognitively demanding tasks. Together, this formulation enables the model to balance fluency with depth, yielding improved performance on benchmarks that require thoughtful, multi-step reasoning.
Parameter updates are performed using the AdamW optimizer with a weight decay of 0.1. The learning rate follows a cosine decay schedule, gradually decreasing from an initial value to 10\% of its peak by the end of training. Stage-specific learning rates are initialized at 1e-5 and 3e-6, respectively, implementing a phased learning rate scheduling approach to balance convergence speed and training stability.

\minisection{Checkpoint Merging} 
Model merging has proven to be an effective technique for integrating diverse task capabilities and enhancing generalization. Existing studies primarily focus on merging heterogeneous models, such as combining SFT models trained with different data mixtures or hyperparameters. Beyond these prior findings, we explore an alternative paradigm: merging homogeneous intermediate checkpoints derived from a single SFT training trajectory. Our approach consolidates checkpoints saved at different stages of the same training run, leveraging their implicit complementary behaviors to improve final model robustness and generalization.

Specifically, we first partition all intermediate checkpoints into distinct groups based on training phases, using epoch indices as indicators. Checkpoints saved within the same epoch are assigned to the same group, ensuring that each group captures unique behavioral characteristics while maintaining temporal coherence. We then perform a weighted aggregation of delta parameters between the SFT checkpoints and the initial base model parameters. Unlike existing methods, we apply a two-layer merging strategy:  1)  \textit{Intra-group merging}: We first aggregate epoch-wise delta parameters within each group; 2) \textit{Inter-group merging}: The resulting epoch-wise deltas are then merged across different groups to obtain the final model parameters.
By incorporating the model merging, we effectively capture and integrate complementary knowledge from different training stages.

Mathematically, we obtain the merged model parameters as follows:
\begin{equation}
\Theta_{merged} = \Theta_{base} + \sum_{k=1}^K \lambda_k \sum_{i=1}^{N_k} \frac{1}{N_k} (\Theta_i^k - \Theta_{base}),
\end{equation}
where $\Theta_{merged}$ denotes the final merged model parameters, $\Theta_{base}$ denotes the initial model parameters of an SFT run, $N_k$ is the number of model checkpoints in the $k\text{-th}$ group, and $\Theta_i^k$ denotes the model parameters of the $i\text{-th}$ checkpoint in the $k\text{-th}$ group.

\minisection{Reinforcement Learning}
We employ the Group Relative Policy Optimization (GRPO) algorithm for the RL policy learning, which is currently a common practice in the post-training RL phase.
A challenge arises when all responses generated for a given prompt receive identical rewards. In such cases, the normalized advantage becomes zero, potentially causing the GRPO objective to degrade into a simple behavior cloning loss, thereby stifling policy exploration. To address this, we introduce a "Zero-Advantage-Mask" mechanism. This mechanism nullifies the loss contribution from samples where the advantage is zero. Consequently, policy updates are driven only by "effective" data exhibiting a clear learning signal (non-zero advantage), promoting more robust exploration and learning.

To provide nuanced and task-specific guidance, we utilize a multi-source reward system that dynamically routes prompts and their generated responses to appropriate evaluators based on task characteristics. This system comprises three key modules:
\begin{itemize}
    \item \textbf{Correctness Rewards:} For tasks with verifiable ground-truth, such as mathematics or coding, correctness-based rewards are assigned. Mathematical problems are assessed by a hybrid system combining rule-based verifiers for standard formats and LLM-based verifiers for more nuanced interpretations. Code responses undergo a multi-stage evaluation: extraction, syntax verification, execution via an online interpreter, and comparison against test cases. Rewards can be structured hierarchically (stage reward) or based on pass rates (continuous reward).
    \item \textbf{Preference Rewards:} For open-domain tasks where ground-truth is unavailable (e.g., creative writing), the system incorporates a preference reward model. This model, typically another LLM trained as a judge, emulates human preferences. Its output scores are normalized before use in GRPO to ensure stability and consistent scaling across diverse prompts.
    \item \textbf{Auxiliary Rewards:} The reward system also includes auxiliary components. A format validator acts as an initial filter, penalizing responses that violate predefined structural requirements. A lightweight repetition penalty, based on string hashing, discourages overly verbose or redundant outputs. These operate orthogonally to the primary reward signals, maintaining output quality without distorting the main learning objectives.
\end{itemize}

To improve training efficiency and effectiveness, we implement a curriculum data mixing strategy. Queries that are consistently too easy (yielding all correct responses) or too difficult (yielding all incorrect responses) produce constant rewards. These offer minimal learning signal for GRPO yet incur computational costs. Our curriculum approach assesses data complexity in a model-aware manner: the current LLM generates multiple diverse responses to a prompt, with complexity determined by the pass rate (for verifiable tasks) or loss (for non-verifiable tasks). Lower pass rates or higher losses indicate greater complexity. Training then proceeds by feeding the model a progressively curated mix of samples with varying complexities, ensuring the model continually receives meaningful reward signals and facilitating more effective and efficient policy updates.

\section{Infrastructure}
\label{sec: Infrastructure}

\subsection{Training System Optimized for Ascend NPUs}
\label{Training}

The advanced accelerate techniques as introduced in Pangu Ultra MoE~\cite{tang2025panguultramoetrain} have been further optimized to deliver improved performance, including refined Hierarchical EP All-to-All Communication with shorter communication volumes, finer-grained operator scheduling and more effective overlapping in Adaptive Pipeline Overlap Mechanism, and additional recomputing and swap modules in memory optimization strategies. These optimizations not only boost the Model FLOPs Utilization (MFU) for Pangu Ultra MoE (details will be published soon), but are also adaptable to Pangu Pro MoE, achieving significant training efficiency improvements with a 35\% relative increase in MFU, as demonstrated in Table~\ref{tab:MFU comparison}. 

\begin{table}[tbp]
    \centering
    \caption{MFU comparison between baseline and after optimization}
    \label{tab:MFU comparison}
    \begin{tabular}{lcccc}
        \toprule
          & Parallelism & Acceleration Strategies & MFU \\
        \midrule
        Baseline &  TP1, EP8, CP8, PP6, VPP4 & - & - \\
        \midrule
         & & Hierarchical EP All-to-All communication & \\
        Optimized & TP8, EP2, CP1, PP5, VPP5 & Adaptive Pipeline Overlap Mechanism & increase by 35\% \\
         & & Fused operators & \\ 
        \bottomrule
    \end{tabular}
\end{table}

To fully exploit these optimizations on Ascend NPUs, the training adopts a carefully tuned parallelism configuration: Tensor Parallelism (TP) = 8, Expert Parallelism (EP) = 2, Pipeline Parallelism (PP) = 5, Virtual Pipeline Parallelism (VPP) = 5, and Context Parallelism (CP) = 1. The sizes of TP and EP are specifically selected to maximize performance under the Hierarchical EP All-to-All Communication scheme. Compared to Pangu Ultra MoE, the EP size is reduced to 2 to minimize EP communication volume when memory capacity allows. The model contains 48 transformer layers, and to achieve better load balancing across pipeline stages, 2 additional no-op layers are appended, increasing the total number of layers to 50. These layers are then evenly partitioned across 5 pipeline stages, with further subdivision using 5 virtual pipeline stages. This 5×5 PP–VPP configuration ensures balanced computation and communication overheads across devices, enhancing the overall scalability and throughput of the training process. Due to the reduced sizes of Pangu Pro MoE and PP-VPP compared with Pangu Ultra MoE~\cite{tang2025panguultramoetrain}, the accumulated activation memory during the warm-up phase is significantly decreased. This diminished memory demand enables stable training without reliance on previously required memory optimizations for Pangu Ultra MoE, including fine-grained recomputation and tensor swapping strategies. Consequently, the training process is further accelerated by eliminating the redundant overhead. Additionally, due to reduced EP communication volumes and the removal of communication overhead in permute recomputation, operator scheduling in Adaptive Pipeline Overlap mechanism achieves full compatibility with Pangu Pro MoE by maximizing the overlap of communication with computation.

Moreover, the proposed Mixture of Grouped Experts (MoGE) architecture in \modelname~ effectively mitigates the computational load imbalance across devices by over 50\%, which is quantified through the reduced maximum disparity in execution time for permute and gmm\_up operators. 

We also optimize operator kernels through architectural refinements. For fundamental operators like Matmul, block size is adjusted during initial data transfers from general memory to L1 cache based on the smaller L0 capacity, which enables earlier L1-to-L0 data transfers, ultimately reducing latency and increasing cube utilization by >10\%. We further optimize the cache strategy based on the feature of the Ascend architecture. By refining the matrix block partitioning method and adjusting the data transfer orders, combined with staggered allocation of computing blocks to mitigate access conflicts, the utilization of HBM bandwidth is significantly enhanced, leading to a 5\% - 10\% improvement in operator performance. Collectively, these systematic optimizations reduce computational redundancy, improve data flow efficiency, and boost training throughput by fully leveraging hardware capabilities.

\subsection{Inference System Optimized for Ascend NPUs}
\label{inference}

\begin{figure}[t]
    \begin{center}
    \centerline{\includegraphics[width=0.6\columnwidth]{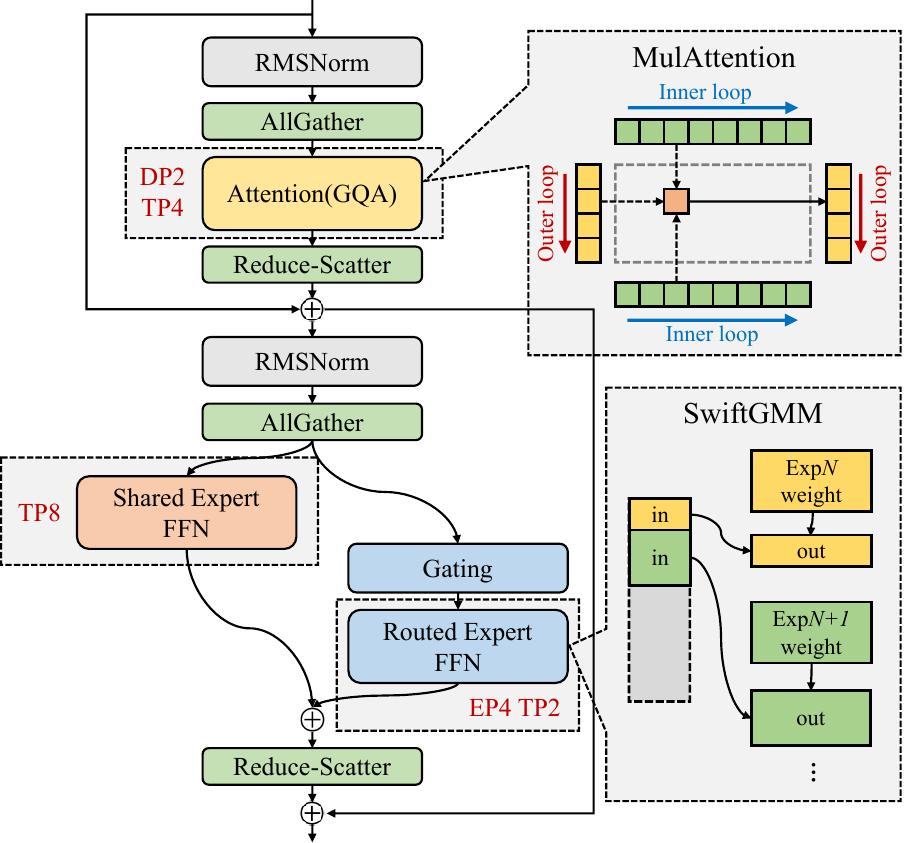}}
    \caption{Overview of the inference system optimization. A $\text{H}^{2}$P strategy is employed to achieve high-efficiency distributed parallel inference across different modules. Additionally, two key fused operators, MulAttention and SwiftGMM, are specifically designed for the Ascend platform to accelerate model inference.}
    \label{infer_overview}
    \end{center}
\end{figure}

\subsubsection{Parallel Optimization}
\label{inference}

\paragraph{Hierarchical \& Hybrid Parallelism}
\modelname~ implements a fused expert system, where sparse expert modules contain 95\% of total parameters while attention mechanisms retain only 5\%. Through systematic analysis of the architectural configuration and hardware specifications of the Ascend computing platform, a hierarchical \& hybrid parallel ($\text{H}^{2}$P) strategy was devised for the deployment of inference, which can eliminate redundant computational operations and inter-process communication bottlenecks to achieve a high computing efficiency, as shown in Figure \ref{infer_overview}.

For attention modules, a hybrid DP2+TP4 parallelism strategy is used to reduce cross-CPU communication overhead on the 300I Duo NPU, where four chips are controlled by one CPU. Requests are grouped along the batch dimension to balance computation between CPU domains. For expert modules, a combination of Tensor Parallelism (TP) and Expert Parallelism (EP) is adopted to address both memory and latency challenges. EP retains full expert matrices for computation efficiency, but causes load imbalance. TP partitions expert matrices for balanced workload, but may reduce efficiency due to suboptimal shapes. A hybrid TP2+EP4 strategy balances these trade-offs based on empirical performance analysis. For shared experts, due to uniform workloads, shared experts use TP8 for dense, efficient, and balanced computation. By applying fine-grained  hybrid parallelism strategies tailored to the model and the Ascend computing platform, the approach achieves optimized inference performance through balanced computation, reduced communication overhead, and efficient hardware utilization.

\paragraph{Communication Strategy}
Building upon the optimized hybrid parallel strategy, we conduct further optimization on the associated communication operations to minimize computational and communication redundancy, as shown in Figure \ref{infer_overview}. 
For the attention module, we adopt the parallel strategy with DP2+TP4. Specifically, inputs are split into two mini-batches where each batch is inferred across four devices.
This configuration originally required one AllReduce and one AllGather operation. To optimize communication efficiency, we replace the AllReduce operation with a Reduce-Scatter operation followed by a AllGather operation, effectively reducing communication data by 50\%.
Furthermore, we strategically reposition the AllGather operator after the RMSNorm operator rather than before it. This adjustment enables parallel execution of RMSNorm computations across devices, achieving a 75\% reduction in computational load through distributed processing.
For the MoE module implementation, we employ a combined TP2+EP4 parallel strategy with shared experts processed through TP8 parallelism. The module's final output integrates results from both routed and shared experts via a global AllReduce operation. To maintain compatibility with the subsequent DP2 attention module while avoiding repartitioning overhead, we decompose the global AllReduce into a global Reduce-Scatter followed by a local AllGather operation across four devices.

\paragraph{Communication-Compute Overlap}
Building on prior work in parallelism and communication optimization, we further reduce communication latency by streamlining the interaction between adjacent computation and communication streams on the Ascend platform. Multi-stream fusion maps computation and communication tasks to separate hardware units, enabling concurrent execution by decoupling data dependencies. In the \modelname~ model, global all-gather and reduce-scatter operations in the Expert component contribute approximately 8\% of the total network latency. To address this, we propose a fine-grained, operator-level compute-communication fusion strategy that decomposes traditionally sequential tasks into interleaved subtasks. By leveraging multi-stream architecture of the Ascend computing platform, we introduce two fused strategies: GMMRS (GroupedMatMul + ReduceScatter) and AGMM (AllGather + MatMul). These enable fine-grained pipeline overlap, improving overall execution efficiency.

\subsubsection{Quantization Compression}
\label{inference}

\paragraph{Expert-Aware Quantization} %

Quantizing Mixture-of-Experts (MoE) models introduces unique challenges due to their sparse and dynamic computation patterns. First, activation outliers in MoE layers exhibit expert-specific distributions, as tokens are routed to distinct subsets of experts. Second, the router’s expert selection mechanism is highly sensitive to quantization-induced logit perturbations. Even minor deviations in gate scores can disrupt the Top-K expert assignment logic, degrading model performance due to misrouted tokens. Third, expert activation sparsity creates calibration bottlenecks: rarely activated experts receive insufficient data coverage during parameter calibration, resulting in inaccurate estimation of quantization parameters and large quantization errors.

To tackle these challenges, we propose a novel expert-aware post-training quantization method. 
Our approach begins with an \textbf{expert-aware smoothing aggregation strategy} designed to suppress activation outliers across MoE experts. By constructing a unified channel-wise smoothing vector that aggregates maximum scaling requirements from both expert weights and router logits, we redistribute outlier magnitudes while preserving mathematical equivalence through parameter fusion with preceding normalization layers. 
For a token vector $\mathbf{x}\in {R}^d$ with $d$ channels, and an MoE layer with $n$ local experts. 
We achieve this through channel-wise maximization over expert-specific and router-specific requirements:
\begin{equation}
\overline{s}_j = \max\left(\underbrace{\max_{i \in [1,n]} \left(\frac{\max(|\mathbf{x}_j|)^{\alpha}}{\max(|\mathbf{W}^i_j|)^{1-\alpha}}\right)}_{\text{Expert requirements}}, ~\underbrace{\frac{\max(|\mathbf{x}_j|)^{\alpha}}{\max(|\mathbf{W}^{\text{gate}}_j|)^{1-\alpha}}}_{\text{Router requirement}} \right),
\end{equation}

where subscript $j$ denotes the $j$-th input channel,  $\alpha$ denotes the migration strength, $\mathbf{W}^i$ denotes the first weight matrix of the $i$-th local expert and $\mathbf{W}^{\text{gate}}$ denotes the weight matrix of the router layer.

To ensure consistent expert selection post-quantization, we introduce \textbf{router logits distribution alignment} through a dual-objective calibration process that minimizes both logit reconstruction error and Kullback-Leibler divergence between full-precision and quantized routing probabilities. This guarantees stable Top-K expert activation despite quantization-induced perturbations.
Finally, we resolve expert-level activation sparsity through \textbf{expert-level calibration data balance}, where underutilized experts receive prioritized sampling from augmented datasets until their activation counts meet parity with computationally derived expectations.

\paragraph{KV Cache Quantization and Sparsity}
KV cache compression~\cite{xiao2023streamingllm,ge2023fastgen,liu2024kivi,li2025kvtuner, yang2025attentionpredictor} is essential to optimize inference infrastructure efficiency, particularly for throughput, context length, and batch size scalability.
Quantization and sparsity techniques can stably mitigate KV preemption while enhancing inference efficiency and overall user experience.
The KVTuner algorithm~\cite{li2025kvtuner} enables an optimized balance between inference efficiency and model accuracy through hardware-friendly mixed-precision quantization. This Ascend-affinitive framework leverages offline profiling and multi-objective optimization to derive Pareto-optimal layer-wise quantization configurations for coarse-grained KV cache segments.
KVTuner's adaptability ensures effective KV cache compression in MoGE architectures by addressing layer-wise sensitivity and dynamic token-expert interactions.

\subsubsection{Kernel Fusion}
\label{inference}

\paragraph{MulAttention} 
With increasing concurrency levels and iterative expansion of sequence lengths, the key-value (KV) cache exhibits linear growth in memory footprint, causing the latency of attention operations to rise to 30\%-50\% of total inference time. 
Consequently, the attention module emerges as a key bottleneck in MoE inference.
Profiling results reveal that KV vector data transfer accounts for approximately 70\% of attention computation time, followed by vector operations and matrix multiplications. Therefore, optimizing both memory access bottlenecks and vector computation constraints within operator pipelines has become a critical challenge in improving the efficiency of attention mechanisms.

\begin{figure}[ht]
    \begin{center}
    \centerline{\includegraphics[width=0.65\columnwidth]{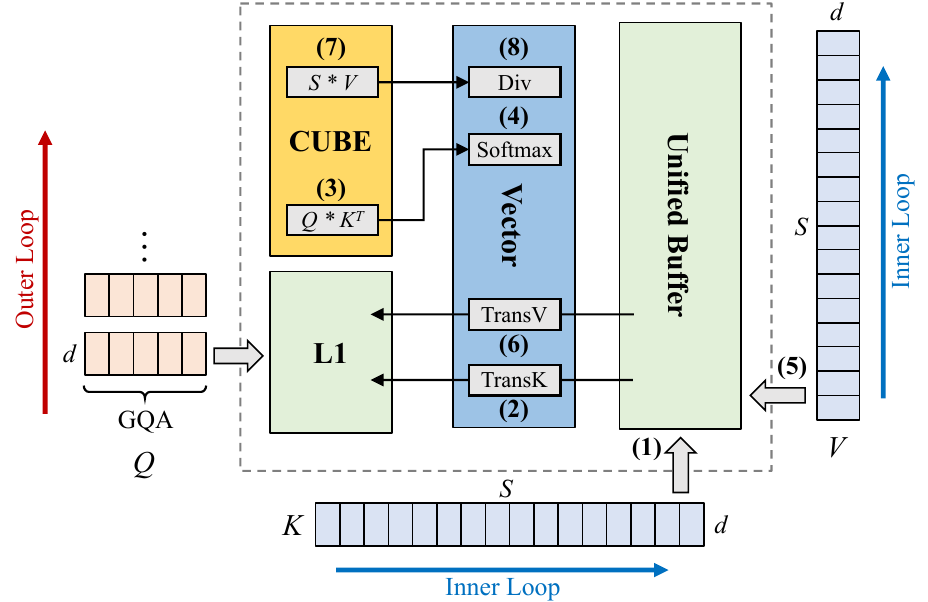}}
    \caption{Computation flow of the MulAttention operator. A large-packet KV transfer strategy is adopted to improve memory bandwidth utilization, as indicated by steps (1)(5). Furthermore, a dual-loop pipeline with a pingpong scheduler is introduced for KV processing to enhance MTE2 utilization, as indicated by steps (2)(3)(4) and (6)(7)(8).}
    \label{infer_mulattention}
    \end{center}
\end{figure}

To address this challenge, we propose \emph{MulAttention}, a fused attention operator optimized for Ascend hardware, specifically designed for the decoding stage in LLM inference. Specifically, MulAttention improves memory bandwidth utilization through a large-packet KV transfer strategy, as illustrated in steps (1) and (5) in Figure \ref{infer_mulattention} . Leveraging the MTE2 transfer unit, KV vectors are block-loaded from global memory (GM) into the Unified Buffer (UB) of the vector computation unit, where a NZ layout transpose is performed simultaneously.

Furthermore, we propose a dual-loop pipeline with a pingpong scheduler for KV processing. By decoupling operations with distinct compute patterns, specifically Cube and Vector operations, into separate loops, we eliminate pipeline bubbles caused by interleaved execution of Key, softmax, and Value computations, as illustrated in steps (2) to (4) and (6) to (8) in Figure \ref{infer_mulattention}. In addition, the use of a ping-pong buffering mechanism allows for the overlapping of KV data prefetching and computation. This overlap effectively hides memory latency and increases the utilization of the MTE2 pipeline to over 89\%. Through these optimizations, MulAttention achieves a 4.5× end-to-end attention speedup, and significantly enhances hardware utilization.

\begin{figure}[t]
    \begin{center}
    \centerline{\includegraphics[width=0.9\columnwidth]{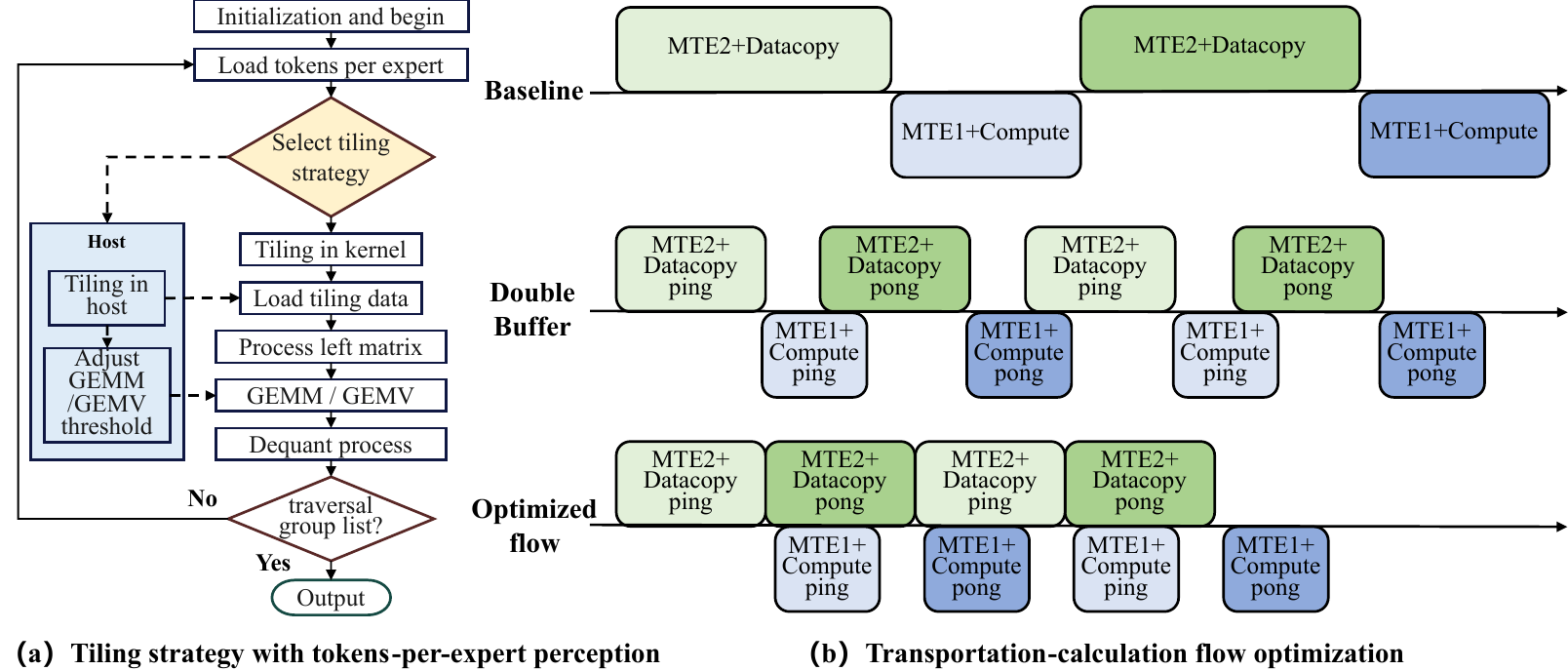}}
    \caption{Overview of SwiftGMM. (a) A tiling cache strategy leverages historical profiling to predict optimal tiling parameters under dynamic workloads. (b) A dual-buffer mechanism overlaps data transfer and computation to maximize MTE2 utilization on the Ascend 300I Duo.}
    \label{infer_swiftgmm}
    \end{center}
\end{figure}

\paragraph{SwiftGMM} 
In high-concurrency scenarios, the GroupMatmul (GMM) operator accounts for over 50\% of end-to-end latency, with dynamic workloads further exacerbating challenges in maintaining computational efficiency. Therefore, we propose \emph{SwiftGMM}, a GMM acceleration technique optimized for the Ascend platform. SwiftGMM introduces a tiling cache strategy tailored to dynamic workloads by leveraging historical profiling data to predict optimal tiling parameters, as illustrated in Figure \ref{infer_swiftgmm} (a), thereby reducing the overhead of frequent recalculations caused by load imbalances. It also dynamically selects between GEMV and GEMM execution modes based on workload intensity to maximize computational throughput.

SwiftGMM further capitalizes on the large L1 cache of the Ascend 300I Duo NPU to load entire matrices in a single pass, substantially minimizing redundant memory transfers. A dual-buffer mechanism is implemented to overlap data movement with computation, thereby enhancing MTE2 pipeline utilization, as shown in Figure \ref{infer_swiftgmm} (b). Experimental evaluations show that SwiftGMM achieves up to 95\% MTE2 utilization, bringing operator performance close to the theoretical upper bound constrained by weight data transfer bandwidth.

\subsubsection{Analysis of Prefill and Decode Stage}
\label{inference}

During the computationally intensive Prefill stage, only the Top-8 experts are activated per token for the MoE architecture, which effectively reduces the model size to an equivalent 16B dense model. This sparse activation mechanism significantly reduces computational cost and communication overhead. Moreover, the adoption of a minimal-card deployment strategy can further enhance computational efficiency in Prefill stage. Compared to popular dense models with sizes of 32B and 72B, \modelname~ achieves much lower latency and higher input throughput under the same hardware conditions.

In the decode stage dominated by memory-intensive operations, \modelname~ maintains low latency within several tens of milliseconds for small batch sizes such as a single batch. For large batch sizes such as 64, the model leverages dimensional compression and depth reduction synergistically with its sparse expert activation paradigm. These features efficiently minimize KV cache memory footprint and inter-node communication overhead, also mitigating computational bottlenecks. As a result, \modelname~ exhibits significantly higher output throughput within an latency such as 100 ms, compared to dense models of similar scale. Detailed experiments on the performance of model inference are presented in Section~\ref{sec:infer_results}.

\section{Experiments and Results}
\label{sec: exp}
In this section, we first present the performance of \modelname~across comprehensive benchmarks, followed by comparative analysis with state-of-the-art models. Then, we analyzed and compared the inference efficiency of \modelname~with dense models. Finally, we investigate the model's expert characteristics through detailed analysis of expert activation patterns during inference to better understand the proposed method.

\subsection{Pre-Trained Model}
Following pre-training on 13T high-quality text tokens, \modelname~achieves significant improvements in linguistic comprehension and reasoning capabilities. To systematically evaluate the model's performance across multiple task dimensions, we conduct a comprehensive evaluation covering three core areas: Chinese language processing, English language understanding, and complex reasoning tasks.
\subsubsection{Evalutaion Benchmarks}
A comprehensive evaluation suite was constructed to assess model capabilities across English, coding, mathematics, and Chinese. Established benchmarks were selected to reflect diverse cognitive skills, with detailed metrics provided in Table~\ref{tab:benchmark-pretrain}.

\textbf{English:}
\begin{itemize}
    \item \textit{General Reasoning}: 
    \textit{Big-Bench-Hard}~\cite{Suzgun2022ChallengingBT},  Massive Multitask Language Understanding(\textit{MMLU})~\cite{mmlu} (multi-subject QA), \textit{MMLU-Pro}~\cite{wang2024mmlu} (complex multi-step reasoning).
    
    \item \textit{Reading Comprehension}: 
    \textit{DROP}~\cite{Dua2019DROPAR} (discrete reasoning over paragraphs), \textit{RACE-M/H}~\cite{Lai2017RACELR} (middle/high school exams).
    
    \item \textit{Commonsense Reasoning}: 
    \textit{HellaSwag}~\cite{Zellers2019HellaSwagCA} (contextual completion), \textit{PIQA}~\cite{Bisk2019PIQARA} (physical commonsense), \textit{WinoGrande}~\cite{sakaguchi2019winogrande} (large-scale Winograd schema).
\end{itemize}
\textbf{Chinese:}
\begin{itemize}
    \item \textit{General Knowledge}: \textit{C-Eval}~\cite{Huang2023CEvalAM} (academic QA), \textit{CMMLU}~\cite{Li2023cmmlu} (linguistic understanding).
    \item \textit{Reading Comprehension}: \textit{CMRC}~\cite{cui-etal-2019-span} (span-extraction RC), \textit{C3}~\cite{sun2019investigating} (multiple-choice RC).
    \item \textit{Cultural \& Contextual}: \textit{CCPM}~\cite{li2021ccpm} (classical poetry matching), \textit{CLUEWSC}~\cite{xu2020clue} (Chinese-language Winograd schema resolution).
\end{itemize}

\textbf{Reasoning:}
\begin{itemize}
    \item \textit{Complex reasoning}: \textit{HumanEval}~\cite{Chen2021EvaluatingLL} (Python function synthesis). \textit{GSM8K}~\cite{gsm8k} (grade school math problem solving), \textit{MATH}~\cite{hendrycks2021measuring} (challenging competition mathematics), \textit{MGSM}~\cite{mgsm} (multilingual mathematical reasoning), \textit{CMath}~\cite{wei2023cmath} (Chinese mathematical problem solving).
\end{itemize}

\subsubsection{Evalutaion Results}
\begin{table}[t]
    \centering
    \footnotesize
    \setlength{\tabcolsep}{4.5pt}
    \caption{Comparison between Pangu Pro MoE  and other representative base models across a diverse set of benchmarks for evaluating language and reasoning skills. \textbf{Bold} values represent the best results in each line.}
    \small
    \begin{tabular}{@{}c l c | ccc | c | c@{}}
    \toprule
    & \multirow{2}{*}{\centering \textbf{Benchmark {\tiny (Metric)}}} & \multirow{2}{*}{\textbf{\# Shots}} & \textbf{Qwen2.5}  & \textbf{GLM4} & \textbf{Gemma3} & \textbf{Llama-4} &  \textbf{Pangu Pro MoE} \\
    & & & \textbf{32B Base} & \textbf{32B Base} & \textbf{27B Base} & \textbf{Scout Base} &  \textbf{Base} \\
    \midrule
    & Architecture & - & Dense  & Dense& Dense& MoE  & MoE \\
    & \# Activated Params  & - & 32B & 32B & 27B & 17B & 16B \\
    & \# Total Params & - & 32B & 32B & 27B & 109B &  72B \\
    \midrule
    \multirow{9}{*}{English} & BBH {\tiny (EM)} & 3-shot & 82.4 & \textbf{83.3} & 77.7
  & 79.5 &  81.2 \\
    & MMLU {\tiny (EM)} & 5-shot & 84.2 &  82.0 & 78.6 &  78.3 &  \textbf{87.4} \\
    & MMLU-Pro {\tiny (EM)} & 5-shot & 58.0 & 55.8 & 50.3 & 50.3 &  \textbf{63.5} \\
    & DROP {\tiny (F1)} & 3-shot & 79.2 & \textbf{86.2} & 77.2 & 77.5 &  81.1\\
    & HellaSwag {\tiny (EM)} & 10-shot & 93.1  & 92.6 & 84.1 &81.9 & \textbf{93.5} \\
    & PIQA {\tiny (EM)} & 0-shot &90.6  & \textbf{91.8} & 83.3 & 88.9 &  89.3 \\
    & WinoGrande {\tiny (EM)} & 5-shot & 80.7 & \textbf{87.1}& 71.9 & 71.4 & 75.6\\
    & RACE-Middle {\tiny (EM)} & 5-shot & 95.3 & 94.0  & 93.8 & 92.3  &  \textbf{95.8} \\
    & RACE-High {\tiny (EM)} & 5-shot &93.0  &92.4 & 90.5 & 88.9  &  \textbf{93.9} \\
    \midrule
    \multirow{6}{*}{Chinese} & CLUEWSC {\tiny (EM)} & 5-shot & \textbf{88.3} & 85.7 &  82.0 & 83.8 &  85.0 \\ 
    & C-Eval {\tiny (EM)} & 5-shot & 87.7 & 84.1 & 69.4 & 74.8  & \textbf{90.6} \\
    & CMMLU {\tiny (EM)} & 5-shot  & 88.9  & 83.8 & 70.4 & 76.8    & \textbf{89.0} \\ 
    & CMRC {\tiny (EM)} & 1-shot  & 66.2  & 76.3 & 70.8 & 67.4   & \textbf{79.8} \\ 
    & C3 {\tiny (EM)} & 0-shot  & \textbf{96.3}  & 94.5 & 62.8 & 94.4  & 95.4\\ 
    & CCPM {\tiny (EM)} & 0-shot  & 86.6  & 87.4 & 69.9  & 81.6  & \textbf{90.3}\\ 
    \midrule
    \multirow{4}{*}{Reasoning} & HumanEval {\tiny (Pass@1)} & 0-shot &57.9 &59.1 & 48.8 & 54.6 & \textbf{63.7} \\
    & MATH {\tiny (EM)} & 4-shot & \textbf{57.9}  &46.7 & 50.0 & 52.7   & 55.9 \\ 
    & CMath {\tiny (EM)} & 3-shot  & \textbf{83.5}  &77.5 & 76.8 & 83.7  & 79.7  \\
    & GSM8K {\tiny (EM)} & 8-shot & 83.0 & 85.4 & 82.6 & 79.2  & \textbf{86.5} \\
    \bottomrule
    \end{tabular}
    \label{tab:benchmark-pretrain}
\end{table}

As demonstrated in Table~\ref{tab:benchmark-pretrain}, Pangu Pro MoE emerges as one of the most competitive architectures across multiple evaluation dimensions. The model establishes state-of-the-art performance in critical English-language benchmarks including MMLU and HellaSwag, while simultaneously dominating most Chinese-language evaluations (C-Eval, C3, and CCPM). Its mathematical reasoning capabilities, as quantified by the GSM8K benchmark, further confirm the architecture's cross-domain competence. Notably, these achievements are attained through a computationally efficient MoE design.
When benchmarked against contemporary base models including Qwen3-32B-base~\cite{yang2025qwen3}, GLM4-32B-base~\cite{glm2024chatglm}, Gemma3-27B-base~\cite{team2025gemma}, and Llama-4-Scout-baset~\cite{metaai_llama4}, Pangu Pro MoE demonstrates consistent performance advantages.

\subsection{Instruct Model}
Benefiting from efficient SFT and RL training, \modelname~ has developed robust instruction-following capabilities and complex reasoning skills. To comprehensively evaluate these capabilities, we conducted systematic assessments on a series of challenging tasks.
\subsubsection{Evalutaion Settings}

\minisection{Evaluation Benchmarks}
We evaluate instructed models on multiple benchmarks across three domains. For general-domain English and Chinese evaluation, we test on: MMLU~\cite{mmlu}, MMLU-Pro~\cite{wang2024mmlu}, MMLU-Redux, DROP~\cite{Dua2019DROPAR}, IF-Eval~\cite{zhou2023instruction}, Arena-Hard~\cite{arenahard2024}, CLUEWSC~\cite{xu2020clue}, C-Eval~\cite{Huang2023CEvalAM}, and CMMLU~\cite{Li2023cmmlu}. To assess reasoning capabilities, we employ code datasets: LiveCodeBench~\cite{jain2024livecodebench}, and MBPP+~\cite{Austin2021ProgramSW}. GPQA-Diamond~\cite{rein2024gpqa} and SuperGPQA~\cite{pteam2025supergpqascalingllmevaluation} is evaluated for scientific reasoning. Mathematical reasoning is evaluated through: AIME 2024~\cite{MAA}, AIME 2025 ~\cite{MAA2025}, MATH-500, and CNMO 2024\footnote[1]{https://www.cms.org.cn/Home/comp/comp/cid/12.html}.

\minisection{Compared Baselines}
We evaluate our instruct model against state-of-the-art models of comparable scale across multiple architectures. The baseline models include dense models (Qwen3-32B~\cite{yang2025qwen3}, GLM4-Z1-32B~\cite{glm2024chatglm}, Gemma3-27B~\cite{team2025gemma}), and MoE models (Llama4-Scout~\cite{metaai_llama4}). For models with public APIs, we conducted evaluations through their official interfaces using standardized configurations. For open-source models without API access, we deployed local instances and performed consistent evaluations under identical settings.

\minisection{Detailed Evaluation Configurations}
To comprehensively assess the performance of the post-training model across diverse datasets, we have adopted a standardized and unified evaluation framework. Specifically, for the LiveCodeBenc, MBPP+, and IF-EVAL datasets, we utilized the evaluation methods officially provided by their respective websites. For the ArenaHard dataset, we employed a referee model to conduct scoring, ensuring a fair and objective assessment. For the remaining datasets, we adopted matching and exact matching techniques to evaluate the model's performance. For LiveCodeBench, we use versions 24.8.1-25.1.1, which cover the data and problem sets between this time period.
Consistent with the original evaluation protocols, we followed the default prompts provided by the dataset creators for all datasets. To ensure the model's capability to handle extensive input and output, we set the maximum input length to 4k tokens and the maximum output length to 28k tokens. This configuration allows us to thoroughly assess the post-training model's ability to process and generate full responses within the specified constraints.

\subsubsection{Evaluation Results} 
\begin{table}[t]
    \centering
    \footnotesize
    \setlength{\tabcolsep}{4.5pt}
    \caption{Comparison between Pangu Pro MoE and other representative models across a diverse set of benchmarks for evaluating language and reasoning skills. \textbf{Bold} values represent the best results in each line.}
    \resizebox{1.0\textwidth}{!}{
    \small
    \begin{tabular}{@{}c l | c c c | c | c@{}}
    \toprule
    & \textbf{Benchmark {\tiny (Metric)}} & \textbf{Qwen3-32B}  & \textbf{GLM-Z1-32B}  & \textbf{Gemma3-27B}  & \textbf{Llama4-Scout} & \textbf{Pangu Pro MoE} \\
    \midrule
    & Architecture & Dense  & Dense & Dense  & MoE&  MoE \\
    & \# Activated Params & 32B & 32B & 27B & 17B &   16B \\
    & \# Total Params & 32B & 32B &  27B &109B &  72B \\
    \midrule
    \multirow{6}{*}{English} & MMLU {\tiny (EM)} &  89.2 & -  & 77.6 & 79.4 &  \textbf{89.3} \\
    & MMLU-Pro {\tiny (EM)} & 78.6 & -  & 67.5 & 73.8 &  \textbf{82.6} \\
    & MMLU-Redux {\tiny (EM)} & 83.2 & \textbf{88.2} & 81.7  & 84.8 &  81.5 \\
    & DROP {\tiny (F1)} & \textbf{91.3} & -& 88.4  & 89.3  & 91.2  \\
    & IF-Eval {\tiny (Prompt Strict)} & 84.3 & 84.5 & 83.4 & 85.2  &  \textbf{85.7} \\
    & Arena-Hard & \textbf{94.7}  & 90.6 & 89.8 & 68.9 & 93.6 \\
    \midrule
    \multirow{3}{*}{Chinese} & CLUEWSC {\tiny (EM)} & 94.6 &  93.6 & 91.3 & 87.6 & \textbf{94.7} \\ 
    & C-Eval {\tiny (EM)} & 89.2 & 82.7  & 65.2  & 77.3 & \textbf{91.1} \\
    & CMMLU {\tiny (EM)}  & 84.6 & -& 65.9  & 70.0 & \textbf{87.1}  \\ 
    \midrule
    \multirow{8}{*}{Reasoning} & LiveCodeBench {\tiny (Pass@1)}   & \textbf{62.6} & 59.1 & 29.7  & 33.1 &  59.6 \\
    & MBPP+ {\tiny (Pass@1)}  & \textbf{82.0} & 75.7  & 73.8  & 73.5 &  80.2 \\
    & GPQA-Diamond {\tiny (Pass@1)}  & 68.2 & 66.1 & 42.4 & 53.5   &  \textbf{73.7} \\
     & SuperGPQA & 49.8 & 52.6 & 35.6 & 45.0 & \textbf{54.8}  \\
    & AIME2024 {\tiny (Pass@1)} & 80.4  & \textbf{80.8} & 26.9  & 29.0 &  79.2   \\
    & AIME2025 {\tiny (Pass@1)} & \textbf{70.9}  & 63.6  & 22.6 & 10.2 & 68.1 \\ 
    & MATH-500 {\tiny (EM)}  & 96.6 & 94.4  & 87.8 & 82.4 &  \textbf{96.8} \\ 
    & CNMO2024 {\tiny (Pass@1)}  & 70.4 & \textbf{71.4} & 40.7  & 20.4 &   70.8\\ 
    \bottomrule
    \end{tabular}
    }
    \label{tab:benchmark-post-training}
\end{table}

\minisection{English Benchmarks} As shown in Table~\ref{tab:benchmark-post-training}, \modelname~demonstrates exceptional performance in English reasoning tasks. On the MMLU-PRO benchmark, which extends MMLU with greater scale and difficulty to rigorously evaluate LLM capabilities. \modelname~significantly outperforms mainstream dense models (including Qwen3-32B, GLM-Z1-32B, and Gemma3-27B) and the MoE-based Llama4-Scout, achieving state-of-the-art results. Notably, \modelname~achieves a competitive score of 91.2 on the DROP reading comprehension task, nearly matching Qwen3-32B's 91.3 score. This demonstrates its semantic understanding in complex English contexts reaches leading levels.

\minisection{Chinese Benchmarks}
 As shown in Table~\ref{tab:benchmark-post-training}, \modelname~maintains comparable expertise in Chinese evaluations. It scores 91.1 on C-Eval (EM), surpassing Qwen3-32B(89.2). For Chinese commonsense reasoning, \modelname~achieves 94.7 on CLUEWSC (EM), outperforming Gemma3-27B (91.3) by 3.4 points while matching Qwen3-32B (94.6). These results validate the model's strong performance in Chinese semantic understanding and commonsense reasoning.

\minisection{Reasoning Benchmarks}
 As shown in Table~\ref{tab:benchmark-post-training}, \modelname~demonstrates superior logical reasoning capabilities with efficient computation. For code generation, it achieves 80.2 on MBPP+, comparable to Qwen3-32B's 82.0. In mathematical reasoning, \modelname~scores 96.8 on MATH-500 (EM), surpassing Qwen3-32B (96.6), and achieves 70.8 on CNMO2024 versus 70.4 for Qwen3-32B. Notably, \modelname~obtains 54.8 on SuperGPQA, significantly outperforming dense models like GLM-Z1-32B (52.6). Remarkably, \modelname~matches 32B-scale state-of-the-art models' reasoning capabilities using only 16B activated parameters. This efficiency stems from the innovative MoGE architecture, which enhances inference speed while maintaining reasoning accuracy.

\begin{table}[t]
\centering
\caption{Inference performance comparison during the Prefill stage on the Ascend 800I A2 NPU. The input sequence length is 2048 with a batch size of 2. TTFT (Time To First Token) measures the forward latency to generate the first token. Throughput is calculated per card.}
\begin{tabular}{@{}c|cc@{}}
\toprule
Model                     & \begin{tabular}[c]{@{}c@{}}TTFT\\ (ms)\end{tabular} & \begin{tabular}[c]{@{}c@{}}Input Throughput\\ (tokens/s)\end{tabular} \\ \midrule
 72BA16B MoE & 424.21                                             & \textbf{4828}                                                                  \\
32B Dense                   & 604.07                                              & 3390                                                                  \\
72B Dense               & 1282.94                                             & 1596                                                                  \\ \bottomrule
\end{tabular}
\label{infer_result_prefill}
\end{table}

\begin{table}[t]
\centering
\caption{Inference performance comparison during the Decode stage on the Ascend 800I A2 NPU. The input sequence length is 2048 tokens. TPOT (Time Per Output Token) represents the forward latency for generating each output token. Throughput is calculated per card. $^*$ represents the acceleration with MTP module and related optimization at a high acceptance rate.}
\begin{tabular}{@{}c|ccc@{}}
\toprule
Model                        & Batch size & \begin{tabular}[c]{@{}c@{}}TPOT\\ (ms)\end{tabular} & \begin{tabular}[c]{@{}c@{}}Output Throughput\\ (tokens/s)\end{tabular} \\ \midrule
\multirow{3}{*}{72BA16B MoE} & 1          & 18.44                                               & 14                                                                     \\
                             & 456        & 99.31                                               & \textbf{1148}                                                          \\
                             & 584        & 95.56                                               & \textbf{1528$^*$}                                                         \\ \midrule
\multirow{2}{*}{32B Dense}   & 1          & 16.26                                               & \textbf{15}                                                            \\
                             & 336        & 86.34                                               & 973                                                                    \\ \midrule
\multirow{2}{*}{72B Dense}   & 1          & 26.88                                               & 9                                                                      \\
                             & 228        & 97.74                                               & 583                                                                    \\ \bottomrule
\end{tabular}
\label{infer_result_decode_new}
\end{table}
\subsection{Inference Efficiency}
\label{sec:infer_results}

\paragraph{Performance on Ascend 800I A2}
\modelname, configured as 72BA16B MoE, exhibits remarkable inference efficiency under the W8A8 quantization on Ascend 800I A2. In the prefill stage, with a batch size of 2 and sequence length of 2k, the model achieves an average input throughput of 4828 tokens/s per card, attaining the lowest prefill latency. Compared to 72B Dense and 32B Dense, this corresponds to performance improvements of 203\% and 42\%, respectively, as shown in Table \ref{infer_result_prefill}. The computational efficiency derives principally from the model's optimized parameter activation pattern, where per-token operational parameters constitute merely 22\% and 50\% of those required by the dense model, thereby substantially mitigating computational demands.

In the Decode stage, four accelerators are deployed for the model. For low concurrency scenarios such as batch size of 1 and the sequence length is 2k, 72BA16B MoE achieves low latency with a weight transfer volume of approximately 16B. For high concurrency scenarios with hundreds of batch size, meeting a typical 100 ms latency constraint, \modelname~ achieves an average output throughput of 1148 tokens/s per card, outperforming 72B Dense and 32B Dense by 97\% and 18\%, as shown in Table \ref{infer_result_decode_new}. Furthermore, the model's output throughput can be increased to 1528 tokens/s per card when incorporating multi-token prediction (MTP) decoding and related optimization.

Profiling results indicate that weight transfer accounts for only 29\% of total latency, with the remaining time primarily spent on KV caching, computation and communication. \modelname~ adopts a smaller hidden dimension of 5120 and fewer layers of 48, leading to reductions in KV cache size and communication volume by approximately 40\%/25\% and 63\%/25\% compared to 72B Dense and 32B Dense, respectively. These structural optimizations fully exploit the computational and memory access advantages brought about by sparse activation and lightweight architecture, significantly enhancing overall inference throughput.

\begin{table}[t]
\centering
\caption{Inference performance comparison during the Prefill and Decode stage on the Ascend 300I Duo NPU. The input sequence length is 2048. Throughput is calculated per card. $^*$ represents the acceleration with MTP module and related optimization at a high acceptance rate.}
\begin{tabular}{@{}c|cccc@{}}
\toprule
Model                        & Stage   & Batch size & \begin{tabular}[c]{@{}c@{}}Latency\\ (ms)\end{tabular} & \begin{tabular}[c]{@{}c@{}}Throughput\\ (tokens/s)\end{tabular} \\ \midrule
\multirow{3}{*}{72BA16B MoE} & Prefill & 2          & 1940.3                                                 & 1055                                                            \\
                             & Decode  & 80         & 99.5                                                   & 201                                                             \\
                             & Decode  & 128        & 99.7                                                   & 321$^*$                                                             \\ \bottomrule
\end{tabular}
\label{infer_result_300I}
\end{table}

\paragraph{Performance on Ascend 300I Duo}
Through deep integration and optimization of 72BA16B MoE with the Ascend platform, the Ascend 300I Duo inference accelerator enables efficient and cost-effective inference for billion-scale MoE models. \modelname~ is quantized under the W8A8 configuration during inference. In the Prefill stage, by employing two Ascend 300I Duo accelerators with a batch size of 2, 72BA16B MoE achieves 1.94s latency for 2k-length input sequences, with an input throughput of 1055 tokens/s per card. In the Decode stage, using the aforementioned $\text{H}^{2}$P deployment on four Ascend 300I Duo accelerators, the system achieves approximately 50 ms latency in low-concurrency scenarios with a single batch and sequence length of 2k, enabling low-latency inference. In high-concurrency settings with batch sizes of 80, it sustains a per-card throughput of 201 tokens/s with 99.5 ms latency, meeting high-throughput demands, as shown in Table \ref{infer_result_300I}. With the acceleration of MTP decoding and related optimization, the model's output throughput can be increased to 321 tokens/s. Through the co-design of the Pangu model and the Ascend platform, we achieve an excellent cost-to-performance ratio for sub-100B model inference on the Ascend 300I Duo.

\paragraph{Performance of Quantization} We performed extensive evaluations of quantization performance across a wide range of benchmarks. As shown in Tables~\ref{tab:quantization}, our approach achieves near-lossless accuracy with W8A8 quantization configurations. Even when quantized with W4A8, the loss of accuracy remains within an acceptable range.

\begin{table}[tbp]
    \centering
    \setlength\abovecaptionskip{0pt}
    \setlength\tabcolsep{7pt}
    \renewcommand\arraystretch{1.1}
    \caption{Quantization accuracy loss under different quantization bit-widths. }
    \begin{tabular}{c|cccc}
        \toprule
         \multirow{2}*{\textbf{Method}} & \multicolumn{4}{c}{\textbf{Accuracy loss}$\downarrow$} \\
         \cmidrule(l){2-5}
         & MATH-500 & Live CodeBench & Arena Hard & GPQA Diamond \\
        \midrule
        W8A8 &  0.80 & 0.37  & 0.60 & 0.51 \\
        W4A8 &  1.20 & 3.68  & 3.40 & 2.53 \\
        \bottomrule
    \end{tabular}
    \vspace{-1.em}
    \label{tab:quantization}
\end{table}

\subsection{Experimental Analysis}

In order to better understand the effectiveness of the MoGE architecture in \modelname~, we systematically examine key characteristics intrinsic to MoE models, such as domain specialization, the dynamics of expert co-activation, intra-group expert distribution and global expert distribution.

\paragraph{Domain Specialization} The pattern of expert specialization serves as a key indicator of the effectiveness of an MoE layer, as it reflects the extent to which experts have successfully learned and internalized knowledge from the data. In this section, we examine the phenomenon of expert specialization across a range of tasks to understand how this pattern varies under different conditions. Our analysis is based on four diverse datasets—C-Eval, MMLU, GSM8K, and HumanEval—which respectively correspond to Chinese language proficiency, English language proficiency, and advanced reasoning abilities in mathematics and programming.

\begin{figure}[h]
	\begin{center}
		\centerline{\includegraphics[width=0.95\columnwidth]{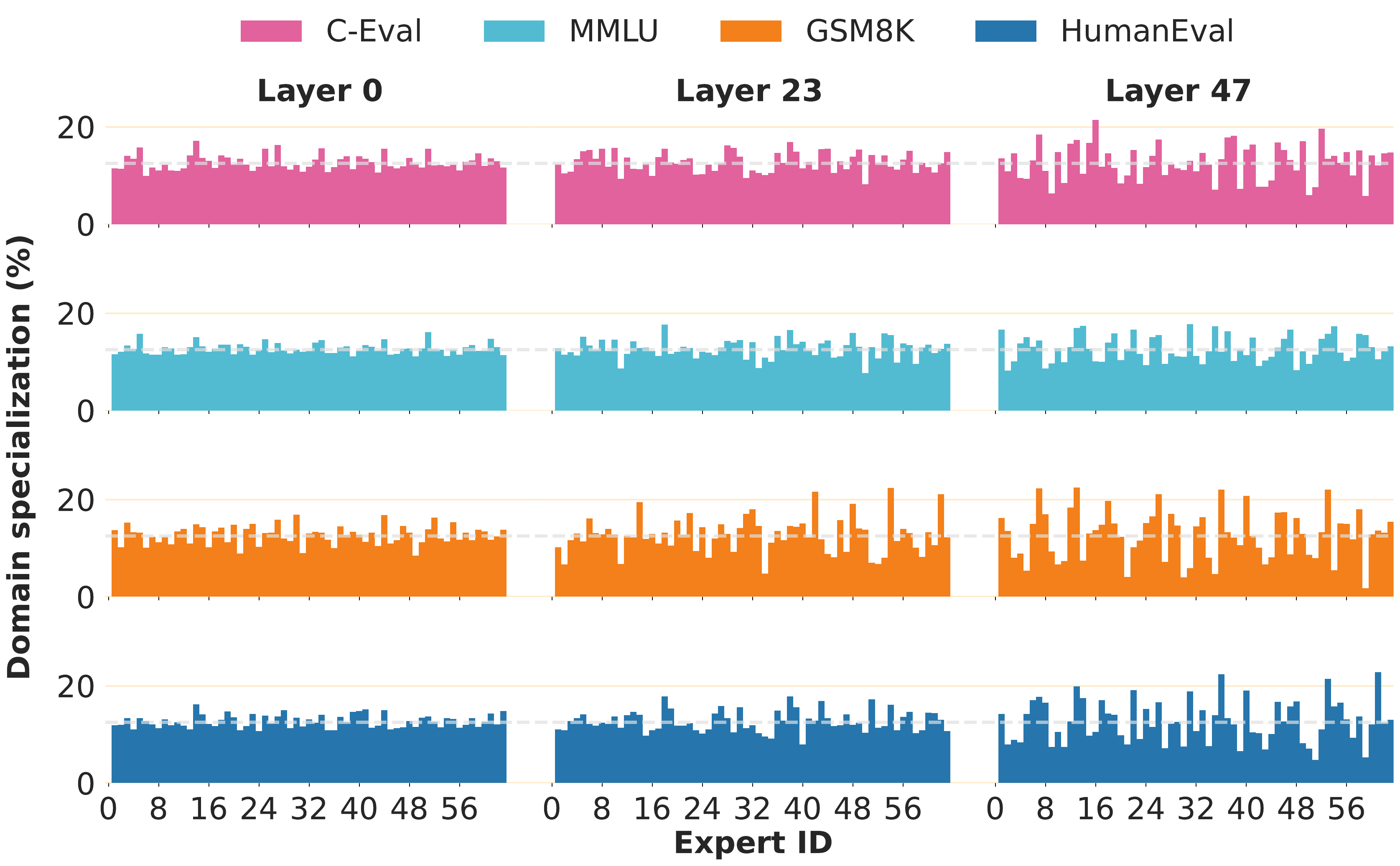}}
		\vspace{-6pt}
		\caption{Expert specialization in \modelname~. Each subplot illustrates the token-to-expert distribution within a specific layer for a given task. Bars represent the proportion of tokens routed to individual experts, normalized by the total number of tokens. \modelname~employs 64 experts per layer, with 8 experts activated for each token, yielding an expected uniform distribution of 12.5$\%$ per expert—indicated by the gray dashed line. The observed distributions, however, deviate substantially from this uniform baseline, highlighting a pronounced degree of expert specialization. This specialization is indicative of effective expert differentiation and contributes to enhanced model training and overall performance.}
		\label{img_routing}
	\end{center}
\end{figure}

As shown in Figure~\ref{img_routing}, we analyze the token-to-expert assignment at three representative layers—shallow, middle, and deep (i.e., Layers 0, 23, and 47). Across different tasks, tokens at the same layer are preferentially routed to different experts, leading to substantial variability in expert specialization. In the shallow layers (Layer 0), the experts exhibit a highly uniform activation pattern. In contrast, experts in deeper layers demonstrate increasing specialization: those in Layer 47 display a higher degree of specialization than those in Layer 23, which in turn surpass those in Layer 0. This progressive trend suggests that expert specialization intensifies with network depth. Furthermore, for tasks that primarily assess general language understanding—such as C-Eval, and MMLU—the distribution of expert activations tends to be more balanced across the expert set. In contrast, for reasoning-intensive tasks such as GSM8K and HumanEval, expert activations exhibit a higher degree of specialization, indicating more selective and task-specific routing behavior.

Our analysis of expert specialization reveals that \modelname~has developed substantial task-specific differentiation among experts, which enhances the model’s representational capacity and contributes to its overall performance.

\paragraph{Expert Co-Activation} 
\begin{figure}[h]
	\begin{center}
		\centerline{\includegraphics[width=0.95\columnwidth]{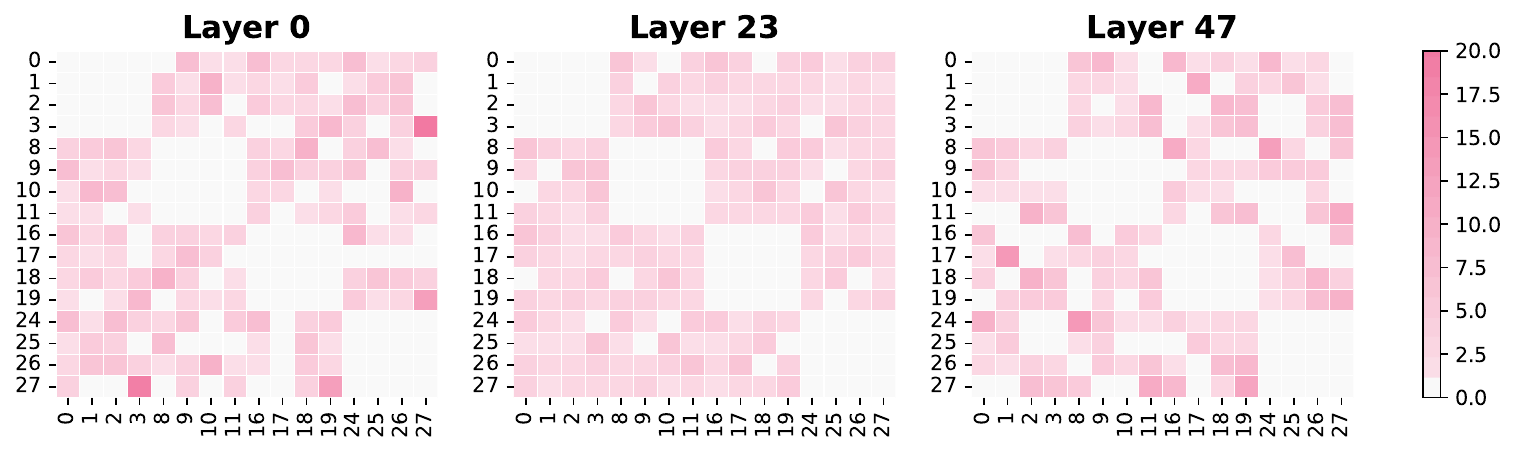}}
		\vspace{-6pt}
		\caption{Expert co-activation across three layers (shallow, middle, and deep), evaluated on a random 0.5$\%$ subset of the C4 validation set. The first four expert groups, each containing four experts, are displayed with their expert IDs on the x- and y-axes. Color intensity reflects the co-activation scores between expert pairs.}
		\label{co_act}
	\end{center}
\end{figure}

To analyze expert interaction behavior, we visualize expert co-activation patterns using a co-activation matrix, where each entry represents the empirical probability that a pair of experts are simultaneously activated for the same input token. Higher co-activation scores indicate stronger correlations in routing decisions, thereby reflecting a higher degree of collaborative behavior among experts. 

To ensure comprehensive coverage across the network, we select three representative layers from different depths—three each from the shallow, middle, and deep stages of the model. As illustrated in Figure~\ref{co_act}, blank regions along the diagonal of the co-activation matrix indicate the absence of co-activation among experts within the same group. This sparsity arises directly from our group-wise routing strategy, which enforces mutual exclusivity in expert selection at the group level, thereby promoting modularization and reducing potential overlap in learned representations.

Additionally, the co-activation scores between experts from different groups remain consistently low across layers, suggesting that inter-group interactions are minimal. This observation supports the hypothesis that our model achieves a low degree of expert redundancy and encourages specialization, where different experts are responsible for distinct aspects of representation learning.

\begin{figure}[h]
	\begin{center}
		\centerline{\includegraphics[width=0.95\columnwidth]{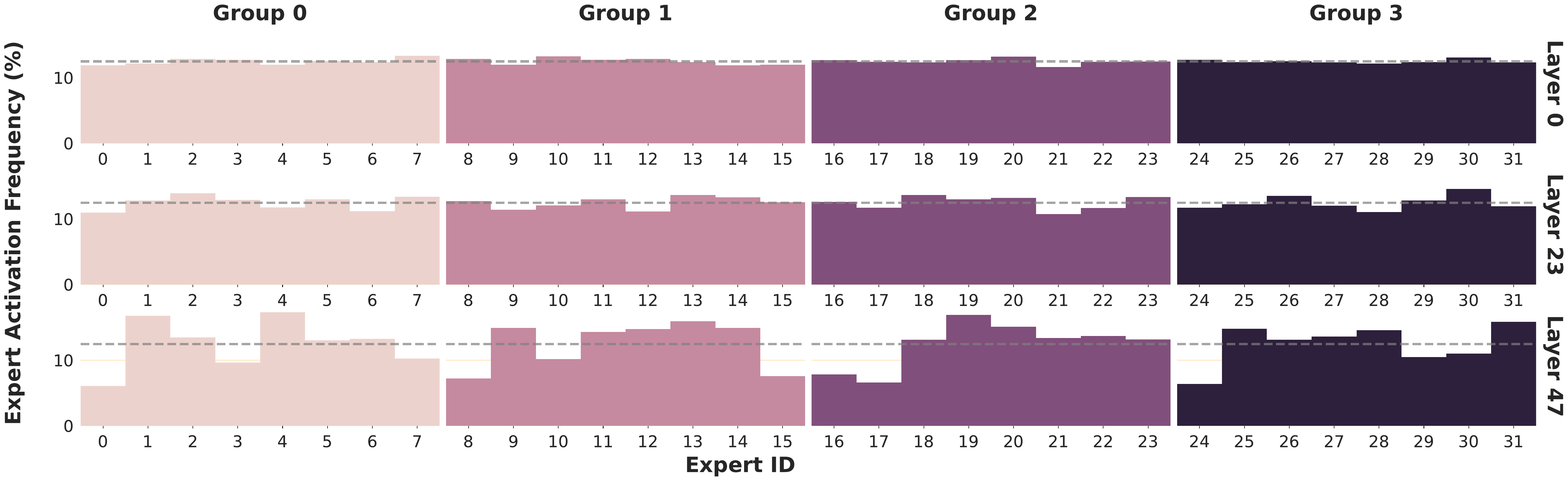}}
		\vspace{-6pt}
		\caption{Intra-group expert distribution in \modelname~. the observed token distributions closely align with this theoretical baseline, demonstrating that \modelname~ effectively maintains balanced utilization across experts within each group. Such balanced activation helps prevent expert underuse or oversaturation, thereby promoting stable training dynamics and maximizing the collective capacity of the expert pool}
		\label{intra_group}
	\end{center}
\end{figure}

\begin{figure}[h]
	\begin{center}
		\centerline{\includegraphics[width=0.95\columnwidth]{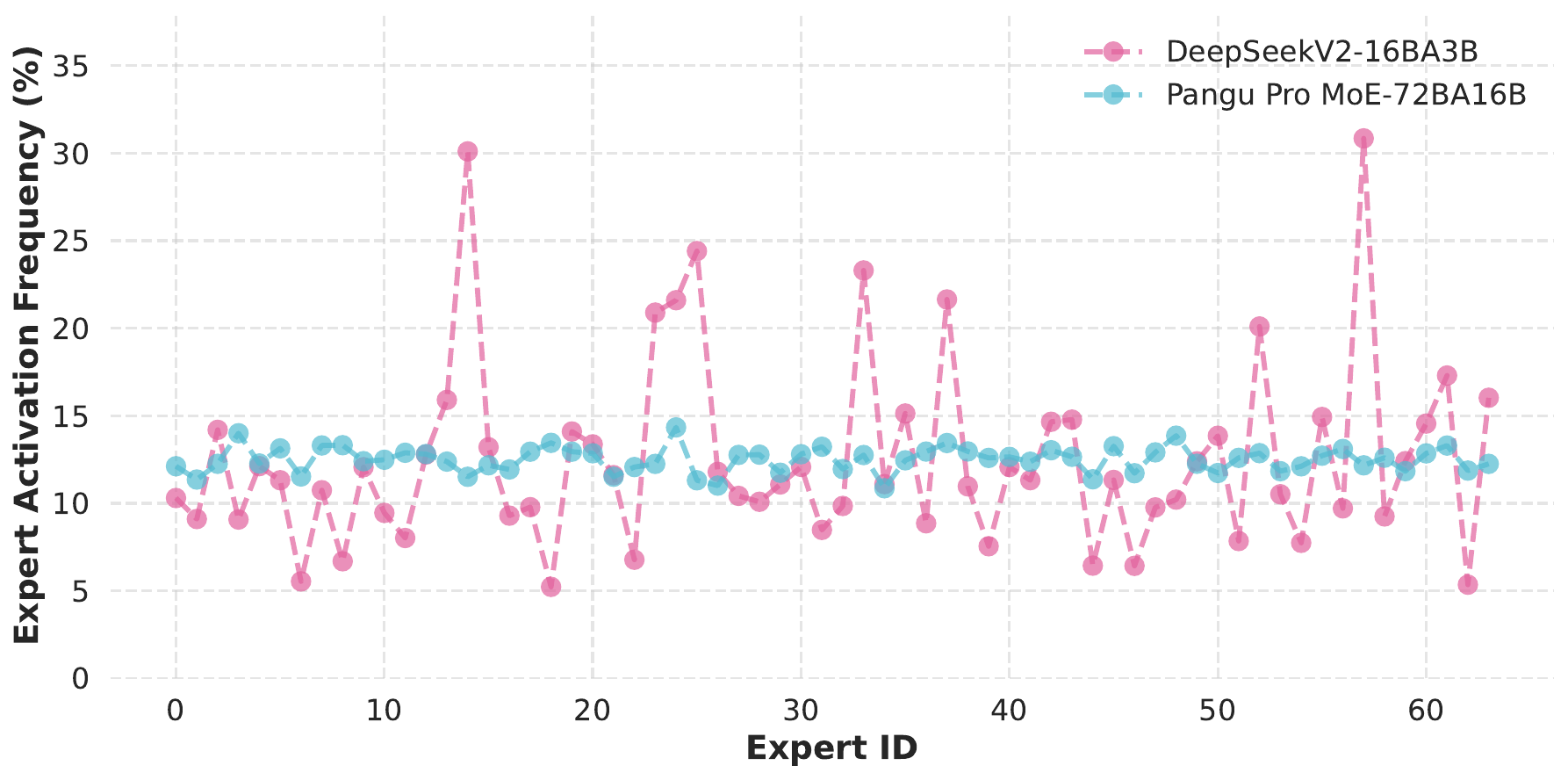}}
		\vspace{-6pt}
		\caption{Global expert distribution on the first MoE layer, evaluated using a random 0.5$\%$ subset of the C4 validation set. Pangu Pro MoE exhibits a more balanced expert utilization, with token proportions closer to the ideal 12.5$\%$ per expert.}
		\label{expert_distribute}
	\end{center}
\end{figure}

Interestingly, we observe a non-uniform trend across layers: co-activation scores are slightly elevated in the shallow and deep layers relative to those in the middle layers. One possible explanation is that the model benefits from broader expert collaboration during early-stage feature extraction—where general-purpose patterns are learned—and during late-stage integration—where diverse signals must be combined for complex task-specific predictions. In contrast, the middle layers may prioritize more fine-grained, isolated processing, leading to greater specialization and reduced inter-expert dependency.

\paragraph{Intra-Group Expert Distribution} 
The design of MoGE enforces the activation of exactly the same experts per group, which inherently promotes balanced expert utilization across groups. To further examine whether this balance also holds within groups, we conduct a detailed analysis of intra-group expert distribution.

As illustrated in Figure~\ref{intra_group}, we visualize the expert activation frequency among experts in the first four groups across three representative layers (shallow, middle, and deep). Overall, the distribution of tokens among intra-group experts appears approximately uniform, with each expert receiving close to 12.5$\%$ of the tokens—consistent with the theoretical average under top-1 activation in a group of 8 experts.
This observation further supports the claim that the MoGE architecture facilitates not only inter-group but also intra-group load balancing, making it inherently friendly to balanced expert utilization.

Notably, we observe slight deviations from perfect uniformity in the deeper layers, where token allocation becomes marginally more skewed. This trend is consistent with the increasing specialization observed in expert routing at greater model depths (Figure~\ref{img_routing}), suggesting that deeper layers may adaptively modulate expert usage to capture more task-specific or abstract representations.

\paragraph{Global Expert Distribution} 
The balance of expert load in MoE architectures remains a critical topic, as more uniform expert activation is generally associated with improved resource efficiency and more stable model behavior. To analyze this aspect, we conduct a comparative analysis between DeepSeek-V2 and Pangu Pro MoE.

As illustrated in the Figure~\ref{expert_distribute}, DeepSeek-V2 exhibits noticeable imbalance, with the most heavily loaded expert processing up to 30$\%$ of the total tokens. In contrast, Pangu Pro MoE demonstrates a nearly uniform distribution across experts, with each expert handling approximately 12.5$\%$ of the tokens—closely aligning with the theoretical ideal.

This balanced activation pattern in Pangu Pro MoE reflects more effective use of the expert capacity and may contribute to enhanced training stability and generalization. The comparison highlights the importance of load balancing in achieving efficient and scalable performance in large-scale MoE models.

\section{Conclusion}
We propose \modelname, a 72B sparse LLM based on the proposed Mixture of Grouped Experts (MoGE) architecture, designed to inherently balance computational workloads across distributed Ascend NPUs. By grouping experts and enforcing balanced token-to-expert assignments, MoGE eliminates device load imbalance issues inherent in conventional MoE, enabling efficient parallel execution during training and inference. Optimized for Ascend 300I Duo and 800I A2 through systematic hardware-aligned co-design, \modelname~activates 16B parameters per token while achieving superior throughput. Extensive experiments validate that \modelname~outperforms leading open-source models such as GLM-Z1-32B and Qwen3-32B, establishing its state-of-the-art capabilities. Our work demonstrates the effectiveness of co-designing sparse architectures with Ascend NPUs for scalable, high-throughput LLM deployment.

\newpage

\bibliographystyle{plain}
\bibliography{ref}

\newpage
\appendix

\section{Contributions and Acknowledgments}

\noindent
\textbf{Core Contributors} \\
Yehui Tang, Xiaosong Li, Fangcheng Liu, Wei Guo, Hang Zhou, Yaoyuan Wang, Kai Han, Xianzhi Yu, Jinpeng Li, Hui Zang, Fei Mi, Xiaojun Meng, Zhicheng Liu, Hanting Chen, Binfan Zheng, Can Chen, Youliang Yan, Ruiming Tang, Peifeng Qin, Xinghao Chen, Dacheng Tao, Yunhe Wang

\textbf{Contributors} \\
An Xiao, Baojun Wang, Bin Wang, Binghan Li, Chenxuan Xiang, Chong Zhu, Dingyu Yong, Dong Li, Dongying Lin, Fan Bai, Fanyi Du, Fisher Yu, Gong Chen, Han Bao, Huan Lin, Huanxin Lin, Huiling Zhen, Jiansheng Wei, Jie Hu, Jing Lei, Jingyong Li, Kaikai Song, Liqun Deng, Miao Rang, Minghui Xu, Nianzu Zheng, Pengfei Xia, Shixiong Kai, Tao Lü, Tianyu Guo, Tiezheng Yu, Wei He, Weizhe Lin, Wenjie Liu, Xing Li, Xiang Lu, Xinduo Liu, Xing Huang, Xu He, Xuan Li, Yao Wang, Yasheng Wang, Ye Tian, Yichun Yin, Yihan Hu, Yinfei Pan, Yixian Ren, Yongbing Huang, Yunsheng Ni, Yuxuan Sun, Zhe Wang, Zheyuan Bai, Zhongqian Fu, Ziyang Zhang, Zongyuan Zhan, Zuming Li

\newpage

\section{Case Study}
To further demonstrate the comprehensive capabilities of \modelname~, we present a series of case studies encompassing both reasoning-intensive tasks, such as mathematics and logical inference, and non-reasoning tasks, including instruction following, general question answering, and AI-generated content (AIGC). These examples underscore the model’s advanced reasoning skills, effective knowledge integration, and superior generation quality relative to baseline models. The key parts that highlight the advantages of \modelname~'s responses have been marked in red.

\hfill

\begin{CJK*}{UTF8}{gkai}
\begin{table*}[ht]
  \centering
  \caption{This AIGC case involves generating a summary of 20 Chinese characters or fewer from an input text. The summary produced by Qwen3-32B lacks detail and omits key content, while \modelname~ preserves more original information within the word limit.}
  \scriptsize
  \resizebox{0.96\textwidth}{!}{
    \begin{tabular}{p{26.5em}l}
    \toprule
    \multicolumn{2}{p{55em}}{\textbf{Question}: 红藕香残玉簟秋。轻解罗裳，独上兰舟。云中谁寄锦书来？雁字回时，月满西楼。花自飘零水自流。一种相思，两处闲愁。此情无计可消除，才下眉头，却上心头。生成20字内摘要} \\
    \midrule
    \multicolumn{2}{p{55em}}{\textbf{Qwen3-32B (thinking mode)}: <think>好的，我现在需要处理用户的一个请求，他们提供了一首李清照的《一剪梅》，然后要求生成一个20字以内的摘要。首先，我得仔细分析这首词的内容和主题。

首先，这首词的上片写的是秋天的景象，红藕香残，玉簟秋，营造出一种凄凉的氛围。接着是“轻解罗裳，独上兰舟”，这里可能表达的是主人公独自泛舟的情景，带有孤独感。然后是“云中谁寄锦书来？雁字回时，月满西楼。”这里用了鸿雁传书的典故，表达对远方亲人的思念，等待书信的心情。

下片“花自飘零水自流”继续用自然景象来比喻时光流逝和无法挽回的离别。“一种相思，两处闲愁”点明了主题，两人分隔两地，共同思念。最后两句“此情无计可消除，才下眉头，却上心头。”强调了相思之苦无法排解，从眉头到心头，情感的深沉。

接下来，我需要将这些元素浓缩成20字以内。要注意关键词：相思、离别、秋景、孤寂、无计可消除。可能需要用简练的词语组合，比如“秋景孤寂，相思难解”或者“离愁别绪，相思缠绵”。但需要确保不超过20字，并且涵盖主要情感和意象。

可能的组合有：“秋景孤寂，相思难消”或者“离愁相思，秋水共长天”。不过要检查字数。例如，“秋景孤寂，相思难解”正好8个字，可能不够。需要更全面。或者“李清照《一剪梅》以秋景抒写离愁，相思难解。”但可能超过20字。需要更精炼。

再想，原词中的“红藕香残”、“兰舟”、“锦书”、“雁字”、“月满西楼”这些意象，但摘要可能不需要具体意象，而是情感和主题。所以重点在“相思”、“离别”、“秋景”、“愁绪”。

最终可能的摘要：“秋景孤寂，相思难解”或者“离愁别绪，相思缠绵”。但可能需要更准确。例如，“秋景孤寂，相思无解”或者“离愁别绪，相思难消”。检查字数，“秋景孤寂，相思无解”是8个字，可能太简略。或者“秋景孤寂，相思难消愁”9个字。或者“李清照《一剪梅》抒写秋日相思离愁。” 这样是13个字，符合要求。或者“秋日相思，离愁难解”8个字，但可能不够全面。需要再调整。

可能最佳的是：“秋景孤寂，相思难解”或者“离愁别绪，相思缠绵”。或者结合季节和情感：“秋日相思，离愁难消”。这样是8个字，符合20字内。或者更准确点：“秋景孤寂，相思无解”或者“秋日相思，离愁难消”。可能后者更好，因为包含了季节和情感，同时“离愁”对应“两处闲愁”，而“难消”对应“无计可消除”。所以最终摘要可以是“秋日相思，离愁难消”正好8个字，或者扩展一点：“秋景孤寂，相思难消”也是8个字。或者“秋日离愁，相思难解”同样8个字。需要确保准确传达原词的意境和情感，同时符合字数限制。</think>

秋日相思，离愁难消
    }\\
    \midrule
    \multicolumn{2}{p{55em}}{\textbf{\modelname~{}}: <think>好的，用户让我给李清照的《一剪梅·红藕香残玉簟秋》生成一个20字以内的摘要。首先，我需要回顾这首词的内容和主题。

这首词是李清照早期的作品，主要表达了她对远方丈夫的思念之情。开头“红藕香残玉簟秋”描绘了秋天的凄凉景象，接着“轻解罗裳，独上兰舟”表现出她独自泛舟的孤独。然后“云中谁寄锦书来？”直接点出对书信的期盼，“雁字回时，月满西楼”进一步渲染了思念的氛围。下阕“花自飘零水自流”用自然景象比喻离别，“一种相思，两处闲愁”说明双方的思念，最后“此情无计可消除，才下眉头，却上心头”强调了情感的无法排遣。

\textcolor{red}{接下来，我需要将这些内容浓缩到20字以内。关键点包括：秋日景象、独处思念、无法消除的相思。要抓住核心情感，即深切的思念和无法排遣的愁绪。可能需要用简洁的词汇组合，比如“秋思”、“独愁”、“相思难消”等。还要注意押韵和节奏，但摘要可能不需要严格押韵，只要准确传达情感即可。}

尝试几个版本：“秋日独愁，相思难消，此情无计可消除。” 但稍长。再简化：“秋思独愁，相思难消，愁上心头。” 还是稍长。可能需要更精炼，比如“秋日独愁，相思难消，此情无计。” 但可能不够完整。或者“秋景凄凉，独上兰舟，相思难消，愁满心头。” 还是字数太多。再调整，去掉具体动作，突出情感：“秋思独愁，两处闲愁，此情难消。” 或者更准确点：“秋日独上兰舟，相思难消，愁满心头。” 但可能超过。需要更精准，可能用“秋思独愁，两处闲愁，此情难消。” 这样符合要求，涵盖了季节、孤独、双方的思念和无法消除的愁绪。</think>

秋思独上兰舟，相思难消，愁满心头。
    } \\
    \bottomrule
    \end{tabular}%
    }
  \label{tab:aigc-case-1}%
\end{table*}%
\end{CJK*}

\hfill

\begin{CJK*}{UTF8}{gkai}
\begin{table*}[ht]
  \centering
  \caption{This is a mathematical case highlighting a key challenge: the ambiguous notation "-(35\%)". This is commonly interpreted in calculators as subtracting 35\% of the preceding number (28.97), but it is not standard mathematical syntax. While Qwen3-32B misinterprets this notation, \modelname~ correctly handles it.}
  \scriptsize
  \resizebox{0.96\textwidth}{!}{
    \begin{tabular}{p{26.5em}l}
    \toprule
    \multicolumn{2}{p{55em}}{\textbf{Question}: 28.97-(35\%)-82*40-58.87=?} \\
    \midrule
    \multicolumn{2}{p{55em}}{\textbf{Qwen3-32B (thinking mode)}: <think>
Okay, let's see. I need to solve this math problem: 28.97 minus 35\%, then minus 82 times 40, then minus 58.87. Hmm, first, I need to figure out what each part means. Let me break it down step by step.

Starting with 28.97 minus 35\%. Wait, when they say 35\%, do they mean 35\% of 28.97? Or is it 35\% of something else? I think in problems like this, when you see a percentage after a number without a specific context, it's usually a percentage of that number. So 35\% of 28.97. Let me confirm that. If I take 35\% of 28.97, that would be 0.35 multiplied by 28.97. Let me calculate that.

First, 0.35 times 28.97. Let me do 0.35 times 28.97. Maybe I can break it down. 0.35 is the same as 35/100 or 7/20. Alternatively, maybe easier to multiply 28.97 by 0.35 directly. Let me try that.

...

So putting it all together:

28.97 - 35\% - 82*40 - 58.87 = -3320.0395.

...

Alternatively, if we keep all decimals until the end and then round, we have -3320.0395, which rounds to -3320.04. Therefore, either way, the answer is -3320.04 when rounded to two decimal places.

Therefore, the final answer is \boxed{-3320.04}.
</think>

To solve the expression:
$$
28.97 - (35\%) - 82 \times 40 - 58.87
$$
we will proceed step by step, interpreting the percentage as 35\% of the preceding number (a common convention in many calculators and practical contexts).

Step 1: Calculate 35\% of 28.97
$$
35\% \text{ of } 28.97 = 0.35 \times 28.97 = 10.1395
$$

...

Step 3: Subtract 58.87
$$
-3261.1695 - 58.87 = -3320.0395
$$
Final Step: Rounding
The original numbers (28.97 and 58.87) are given to two decimal places, and the final result is expected to be in a similar format. Rounding **-3320.0395** to two decimal places:
$$
-3320.0395 \approx -3320.04
$$
Final Answer:
$$
\boxed{-3320.04}
$$
    }\\
    \midrule
    \multicolumn{2}{p{55em}}{\textbf{\modelname~{}}: <think>
Okay, let's tackle this problem step by step. The expression is 28.97 minus 35\% minus 82 times 40 minus 58.87. Hmm, order of operations is crucial here. Remembering PEMDAS—Parentheses, Exponents, Multiplication and Division, Addition and Subtraction. So first, I need to handle the percentage and the multiplication before dealing with the subtractions.

Starting with the original expression: 28.97 - 35\% - 82 * 40 - 58.87. Let's break it down term by term. The percentage sign can be a bit confusing. Is the 35\% referring to 35\% of the previous number, which is 28.97, or is it 35\% of the entire expression? The way it's written, without parentheses, I think it's just 35\% as a standalone value. But percentages can be tricky. Usually, if it's written like this, 35\% would be equivalent to 0.35 in decimal form. So maybe I need to convert that percentage to a decimal first.

\textcolor{red}{So converting 35\% to decimal is 0.35. Now, does that mean we subtract 0.35 from 28.97? Let me check. If the problem was 28.97 - 35\%, then yes, it's 28.97 minus 0.35. But sometimes, in certain contexts, people might interpret 35\% of 28.97, which would be 0.35 * 28.97. But since there's no parentheses indicating that, I think the correct interpretation is just subtracting 0.35.}

...

Next, subtract 3280 from 28.62. That would be 28.62 - 3280. Since 3280 is much larger than 28.62, this will result in a negative number. Let's compute 3280 - 28.62 first, which is 3251.38, and then apply the negative sign. So 28.62 - 3280 = -3251.38. Now the expression is -3251.38 - 58.87.

Finally, subtract 58.87 from -3251.38. That's like adding the two negatives. So -3251.38 - 58.87 = -(3251.38 + 58.87). Let's add those two numbers. 3251.38 + 58.87. Adding the whole numbers: 3251 + 58 = 3309. Then the decimals: 0.38 + 0.87 = 1.25. So total is 3309 + 1.25 = 3310.25. Therefore, the result is -3310.25.

Wait, but let me double-check the steps to make sure I didn't make a mistake. Starting with 28.97 - 0.35 = 28.62. Then 28.62 - 3280 = -3251.38. Then -3251.38 - 58.87 = -3310.25. That seems right. Let me verify the addition again: 3251.38 + 58.87. 3251 + 58 is 3309, 0.38 + 0.87 is 1.25, so 3309 + 1.25 is indeed 3310.25. So the final result is -3310.25.

...

Then subtract 3280: -30.25 -3280 = -3310.25

Yes, same result. So that seems consistent.

Therefore, I believe the correct answer is -3310.25.

**Final Answer**
The result of the expression is \boxed{-3310.25}.</think>

To solve the expression \(28.97 - 35\% - 82 \times 40 - 58.87\), we follow these steps:

1. Convert the percentage to a decimal:
   35\% = 0.35

2. Perform the multiplication:
   $82 \times 40 = 3280$

3. Substitute the converted percentage and multiplication result back into the expression:
   28.97 - 0.35 - 3280 - 58.87

4. Perform the subtractions from left to right:
   - Subtract \(0.35\) from \(28.97\):
     28.97 - 0.35 = 28.62
    
   - Subtract \(3280\) from \(28.62\):
     28.62 - 3280 = -3251.38

   - Subtract \(58.87\) from \(-3251.38\):
     -3251.38 - 58.87 = -3310.25

Thus, the result of the expression is \(\boxed{-3310.25}\).
    } \\
    \bottomrule
    \end{tabular}%
    }
  \label{tab:aigc-case-1}%
\end{table*}%
\end{CJK*}

\begin{CJK*}{UTF8}{gkai}
\begin{table*}[ht]
  \centering
  \caption{This is an example of logical reasoning. When addressing abstract and complex problems, \modelname~ employs multiple strategies to verify and refine its reasoning, ultimately arriving at the correct solution.}
  \scriptsize
  \resizebox{0.96\textwidth}{!}{
    \begin{tabular}{p{26.5em}l}
    \toprule
    \multicolumn{2}{p{55em}}{\textbf{Question}: 如果昨天是明天就好了，那么今天就是周五了，请问今天周几？} \\
    \midrule
    \multicolumn{2}{p{55em}}{\textbf{Qwen3-32B (thinking mode)}: <think>嗯，这个问题看起来有点绕，不过让我慢慢来想。题目是说：“如果昨天是明天就好了，那么今天就是周五了，请问今天周几？”首先，我需要理清楚这个条件句的逻辑关系。这里的“如果昨天是明天”是什么意思呢？可能是指把“昨天”当作“明天”来处理，或者说在时间上进行某种调整，使得原本的“昨天”变成“明天”，从而导致“今天”变成周五。

首先，我需要明确几个时间点的关系。通常来说，今天、昨天和明天的关系是固定的：昨天是今天之前的那一天，明天是今天之后的那一天。但在这个问题中，假设“昨天”变成了“明天”，也就是说，原本的昨天被替换成了明天，这可能意味着时间上发生了某种移动或者调整

...

或者，可能应该这样考虑：假设在某种情况下，把“昨天”当作“明天”，那么原来的今天应该被调整到某个位置，使得当昨天变成明天时，新的今天就是周五。例如，如果实际的今天是X，那么当把X-1（昨天）当作X+1（明天）时，那么新的今天应该是X-1 -1 = X-2，因为如果原来的昨天（X-1）现在是明天，那么新的今天应该是X-1的前一天，也就是X-2。而根据题目，此时的X-2就是周五，所以X = 周日。

是的，这似乎合理。所以答案应该是周日。不过让我再举一个例子验证。比如，如果今天是周日，那么昨天是周六。如果把周六当作明天，那么新的今天应该是周五，而题目说这时候的今天就是周五，所以正确。因此，实际的今天应该是周日。

不过也有可能我哪里弄错了，比如是否应该把“如果昨天是明天”理解为将今天向后推两天？比如，如果今天是周五，那么昨天是周四，但题目说如果昨天是明天，也就是周四变成明天，那么新的今天应该是周四的前一天，也就是周三？但这样的话，新的今天应该是周三，而题目说这时候的今天是周五，所以这显然不对。所以这种情况下，可能我的初始分析是对的，即实际的今天应该是周日。</think>

**题目解析**
题目是：“如果昨天是明天就好了，那么今天就是周五了。”  
我们需要找出**实际的今天是星期几**，使得当“昨天”被当作“明天”时，新的“今天”是周五。

---

**设定变量**
设实际的今天是 **X**（星期几）。  
- 正常情况下，**昨天**是 **X - 1**，**明天**是 **X + 1**。  
- 题目中说“如果昨天是明天”，即把 **X - 1** 视为 **X + 1**。  
- 这意味着，时间发生了偏移：**X - 1 = X + 1**，即时间整体向前移动了 **2天**。  

因此，当“昨天是明天”时，**新的今天**应该是 **X - 2**（因为时间提前了两天）。  
根据题目，此时新的今天是**周五**，即：
$
X - 2 = \text{周五}
$

解得：

$$
X = \text{周日}
$$

---

**验证**
- 实际的今天是**周日**。  
- 正常的昨天是**周六**，明天是**周一**。  
- 如果“昨天是明天”，即把**周六**当作**明天**，那么新的今天就是**周六的前一天**，即**周五**。  
- 这与题目描述一致。

---

**结论**
**今天是周日。**
    }\\
    \midrule
    \multicolumn{2}{p{55em}}{\textbf{\modelname~{}}: <think>嗯，这个问题看起来有点挑战性，不过让我慢慢来想一想。题目是说：“如果昨天是明天就好了，那么今天就是周五了，请问今天周几？”首先，我需要理解这句话的结构，可能涉及到时间的相对转换。

首先，我需要明确题目中的各个时间点之间的关系。题目中的“如果昨天是明天”这个条件，听起来有点矛盾，因为昨天和明天是两个不同的时间方向。可能需要通过代数的方法来设定变量，然后建立方程来解决。

假设今天的实际日期是X。那么根据题目，如果“昨天”（也就是X-1）变成“明天”，那么今天就会变成周五。也就是说，当把“昨天”替换为“明天”时，新的今天应该是周五。这时候需要明确这里的替换是如何影响日期的。

...

具体来说，假设原来的昨天是Y，那么如果Y变成明天的话，也就是Y = X' + 1，其中X'是调整后的今天（即周五）。因此，原来的昨天Y等于调整后的明天，也就是周五 + 1 = 周六。所以原来的昨天是周六，那么原来的今天就是周六 + 1 = 周日。因此，今天应该是周日？

不过这个结论是否正确呢？让我再检查一遍。

...

另一种情况是，把原来的明天（X+1）当作昨天来看待，此时的今天就是周五。因此，原来的X+1被视为昨天，那么今天就是周五，所以原来的X+1=周五-1=周四，因此X=周四-1=周三。这时候今天就是周三。

因此，根据不同的解释方式，答案可能是周日或周三。但需要确定哪一种解释是正确的。

...

\textcolor{red}{现在，我需要验证这两种解法哪一种正确。}

\textcolor{red}{例如，如果今天是周三，那么实际的昨天是周二，实际的明天是周四。题目中说，如果昨天（假设中的昨天）是明天（实际中的明天），即如果假设中的昨天是周四，那么假设中的今天就是周四 +1 = 周五，这符合题意。因此，当实际今天是周三时，假设中的昨天是周四（即实际中的明天），那么假设中的今天就是周五，因此符合题目的条件。}

\textcolor{red}{因此，正确答案应该是周三。}

\textcolor{red}{而之前的周日解法可能是误解了题意的方向，即把实际中的昨天当作假设中的明天，从而导致不同的结果。因此，正确的答案应该是周三。}

总结一下：

当题目说“如果昨天是明天就好了，那么今天就是周五了”，这里的“昨天”指的是假设中的昨天，而“明天”指的是实际中的明天。因此，假设中的昨天 = 实际中的明天，而假设中的今天就是周五。因此：

假设中的昨天 = 实际中的明天 → 周五 -1 = X +1 → X = 周三

因此，今天实际是周三。</think> 

今天实际的日期是周三。以下是推理过程：

1. **设定变量**：设实际今天为 \( X \)。

2. **条件分析**：题目假设“如果昨天是明天”，即假设中的昨天等于实际中的明天。此时假设中的今天为周五。

3. **建立方程**：  
   假设中的昨天 = 实际中的明天 → \( 周五 - 1 = X + 1 \)。

4. **求解**：  
   \( 周四 = X + 1 \) → \( X = 周三 \)。

**验证**：  
- 若实际今天是周三，实际昨天是周二，实际明天是周四。  
- 根据题意，假设“昨天（假设中的）是明天（实际中的）”，即假设中的昨天为周四，则假设中的今天为 \( 周四 + 1 = 周五 \)，与题目条件一致。

**答案**：今天周三。
    } \\
    \bottomrule
    \end{tabular}%
    }
  \label{tab:aigc-case-1}%
\end{table*}%
\end{CJK*}

\begin{CJK*}{UTF8}{gkai}
\begin{table*}[ht]
  \centering
  \caption{This is an instruction following case. There are two instructions on the question. One is that the answer must contain more than six "!". The other is to repeat the question and not to repeat the instruction itself. \modelname~ completes both instructions perfectly, but Qwen3-32B does not meet the second one.}
  \scriptsize
  \resizebox{0.96\textwidth}{!}{
    \begin{tabular}{p{26.5em}l}
    \toprule
    \multicolumn{2}{p{55em}}{\textbf{Question}: Write a short startup pitch for a new kind of ice cream called "Sunnis ice cream". The ice cream should be gentle on the stomach. Contain 6 or more exclamation marks "!" in your response.
    
    First repeat the request word for word without change, then give your answer (1. do not say any words or characters before repeating the request; 2. the request you need to repeat does not include this sentence)} \\
    \midrule
    \multicolumn{2}{p{55em}}{\textbf{Qwen3-32B (thinking mode)}: <think>Okay, the user wants me to create a startup pitch for "Sunnis ice cream" that's gentle on the stomach and includes six or more exclamation marks. Let me start by understanding the key points. The ice cream needs to be stomach-friendly, so maybe it's lactose-free or uses alternative ingredients. I should highlight that aspect first.

Next, the exclamation marks. I need to make sure there are at least six. They can be used in the product benefits, the call to action, and maybe in the tagline. Let me brainstorm some phrases. Words like "delicious," "gentle," "refreshing," and "indulge" come to mind.

I should structure the pitch with an attention-grabbing opening, mention the unique selling points (like being easy on the stomach), list the flavors, and end with a strong call to action. Maybe add some emojis for visual appeal. Let me check if I have six exclamation points. Let's count: "Introducing..." has one, "crafted..." has two, "No more..." has three, "Try..." has four, "Join..." has five, and "Sunnis..." has six. Wait, that's only six. Maybe add another one in the tagline. Let me revise that.

Also, the user wants the request repeated exactly first. I need to make sure I don't add any extra text before that. Alright, let me put it all together now.</think>

Write a short startup pitch for a new kind of ice cream called "Sunnis ice cream". The ice cream should be gentle on the stomach. Contain 6 or more exclamation marks "!" in your response.

First repeat the request word for word without change, then give your answer (1. do not say any words or characters before repeating the request; 2. the request you need to repeat does not include this sentence)

Introducing Sunnis Ice Cream—the revolutionary treat crafted for sensitive tummies and sweet cravings! Our ice cream is made with a unique blend of probiotics, plant-based ingredients, and zero artificial additives, ensuring every scoop is as gentle on your stomach as it is delightful to your taste buds! No more bloating, no more discomfort—just creamy, dreamy goodness that melts into pure joy! With flavors like Minty Probiotic Bliss, Caramel Calm \& Cool, and Vanilla Soothe \& Sparkle, there’s something for everyone to savor! Whether you’re a lifelong ice cream lover or someone who thought they had to say goodbye to their favorite dessert, Sunnis is here to save the day (and your digestion)! Try it today and feel the difference—your stomach will thank you, your friends will rave, and your summer will be legendary! Join the \#SunnisScoop movement and discover why gentle ice cream is the future of frozen fun!}\\
    \midrule
    \multicolumn{2}{p{55em}}{\textbf{\modelname~{}}: <think>Okay, the user wants me to write a startup pitch for "Sunnis Ice Cream" that's gentle on the stomach and includes at least six exclamation marks. Let me start by understanding the key points here.

First, the product is ice cream, but it needs to be easy on the stomach. That probably means it's lower in lactose, maybe uses alternative milk sources like almond, coconut, or oat milk. Or perhaps it's sorbet? Also, could be fewer artificial additives that often cause digestive issues. I should highlight that it's digestible without causing discomfort.

Next, the pitch needs to be engaging and enthusiastic. Exclamation marks are crucial here—six or more. I need to make sure the tone is upbeat and exciting. Maybe use phrases like "Say goodbye to tummy troubles!" or "Indulge guilt-free!"

The target audience isn't just ice cream lovers but also those with sensitive stomachs, lactose intolerance, or dietary restrictions. Emphasize inclusivity. Words like "everyone," "all," "without compromise" could work.

Structure-wise, start with a catchy opener. Maybe a question or a bold statement. Then introduce the product, explain the unique selling points (gentle on the stomach, delicious flavors), mention the ingredients, and end with a call to action. Sprinkle those exclamation marks throughout to keep the energy high.

Wait, the user specified "short" pitch. So keep it concise but impactful. Maybe three to four sentences. Let me draft a sample:

"Introducing Sunnis Ice Cream—finally, a treat that’s as kind to your tummy as it is to your taste buds!!! Made with ultra-gentle ingredients and no harsh additives, enjoy every creamy bite without the dreaded post-scoop regret!! Whether you’re lactose-sensitive, vegan, or just craving peace of mind, Sunnis delivers joy in every spoonful!! Flavors so rich, so smooth, you’ll wonder why other ice creams even exist!! Join the revolution—where indulgence meets gut-friendly goodness!!!!"

Let me check the exclamation count. There are six in total. Wait, the first line has three, then two, then two, then two, and the last one has three. That's more than six. Also, the message covers the key points: gentle ingredients, no additives, targets specific groups, mentions flavors, and a call to action. Sounds good. Make sure the brand name is correctly spelled as "Sunnis" and the message flows naturally with excitement.</think>

\textcolor{red}{Write a short startup pitch for a new kind of ice cream called "Sunnis ice cream". The ice cream should be gentle on the stomach. Contain 6 or more exclamation marks "!" in your response.}

Presenting **Sunnis Ice Cream**—the revolutionary treat that’s **so gentle on your stomach**, you’ll forget you’re eating ice cream!!! Crafted with ultra-easy-to-digest ingredients and zero harsh additives, Sunnis is here to save your tummy *and* your taste buds!!! Say goodbye to bloating, guilt, or post-scoop regrets—just pure, creamy happiness!!! Whether you’re lactose-sensitive, vegan, or just craving peace of mind, Sunnis delivers indulgence without the ouch!!! Flavors like Coconut Bliss, Chocolate Zen, and Vanilla Harmony are **creamy**, **dreamy**, and **stomach-approved**!!! Join the ice cream revolution—finally, a scoop that cares!!!
    } \\
    \bottomrule
    \end{tabular}%
    }
  \label{tab:aigc-case-1}%
\end{table*}%
\end{CJK*}

\begin{CJK*}{UTF8}{gkai}
\begin{table*}[ht]
  \centering
  \caption{This case focuses on Chinese cultural knowledge. The hand ceremony does not carry the meaning of "right hand outside = friendly," which is a broader factual misunderstanding addressed by Qwen3-32B. \modelname~'s response aligns more closely with authentic Chinese cultural practices.}
  \scriptsize
  \resizebox{0.96\textwidth}{!}{
    \begin{tabular}{p{26.5em}l}
    \toprule
    \multicolumn{2}{p{55em}}{\textbf{Question}: 行拱手礼的时候应该哪只手在外？} \\
    \midrule
    \multicolumn{2}{p{55em}}{\textbf{Qwen3-32B (thinking mode)}: <think>嗯，用户问的是行拱手礼的时候应该哪只手在外。首先，我需要确定拱手礼的基本礼仪和传统。拱手礼是中国传统的见面礼节，通常用于表示尊敬或友好。根据我所知道的，拱手礼有两种主要形式：左手在外和右手在外，分别对应不同的场合。

接下来，我需要回忆具体的规则。一般来说，左手在外是用于对长辈、上级或者表示尊敬的情况，而右手在外则用于平辈或朋友之间。这可能涉及到左右手的不同象征意义，比如左为上，右为次。不过，我需要确认这个说法是否准确，是否存在例外情况。

然后，用户可能是在学习中国传统文化，或者需要在实际场合中使用拱手礼，比如婚礼、传统节日活动等。他们可能希望了解正确的礼仪，以避免失礼。因此，回答时需要明确不同情况下的正确做法，并给出例子，帮助用户理解。

另外，用户可能没有明确说明场合，所以需要涵盖不同情境下的应用，比如日常问候、正式场合、不同性别之间的使用等。还要注意是否有地区差异，比如南方和北方在拱手礼上的习惯是否一致。

还需要考虑是否有现代的变化，比如在当代社会中，拱手礼是否仍然普遍使用，或者是否有其他变体。同时，是否需要提到拱手礼的其他细节，比如动作的幅度、是否鞠躬等，但用户的问题集中在“哪只手在外”，所以可能不需要深入其他细节。

最后，要确保信息准确，引用可靠的来源，比如传统礼仪书籍或权威的文化资料。如果有不确定的地方，应该建议用户进一步查阅或咨询专业人士，但根据现有知识，尽量给出清晰明确的答案。</think> 

在中国传统礼仪中，**行拱手礼时，哪只手在外**取决于具体场合和对象，以下是详细说明：

**1. 基本规则**

- **左手在外**（左手在前，右手在后）：  
  用于对**长辈、上级、尊贵者**行礼，表示尊重和谦逊。  
  - **适用场景**：向父母、老师、领导、长者等表达敬意时。  
  - **文化依据**：古代以左为尊（如“左丞右相”中左为高位），但需注意现代礼仪中部分场合已简化。

- **右手在外**（右手在前，左手在后）：  
  用于**平辈、朋友、晚辈**之间，表示平等或友好。  
  - **适用场景**：与同龄人、同事、朋友打招呼时。  
  - **特殊说明**：若对方为女性，男性行礼时右手在外更显礼貌（避免左手在外可能的误解）。

**2. 补充细节**

- **动作要领**：双手合抱于胸前，左手在外时指尖向右，右手在外时指尖向左，动作自然舒展，可配合微微鞠躬。  
- **现代简化**：日常生活中，拱手礼常简化为双手合抱（不区分左右），仅通过动作幅度和态度表达诚意。  
- **地区差异**：部分地区（如南方）更注重左右手之分，北方则相对灵活。

**3. 常见误区**

- **错误1**：认为“左手永远在内”或“右手永远在内”，需根据对象调整。  
- **错误2**：对女性行礼时强行左手在外，可能引发尴尬，建议右手在外更稳妥。  

**4. 实际应用示例**

- **对长辈**：双手拱起，左手在外，略低头致意。  
- **对平辈**：双手拱起，右手在外，面带微笑。  
- **正式场合**（如婚礼、祭祀）：严格遵循左手在外的传统礼仪。  

**总结**  

**左手在外 = 尊敬；右手在外 = 友好**。若不确定场合，可选择双手自然合抱（不刻意区分左右），以真诚的态度为主。
    }\\
    \midrule
    \multicolumn{2}{p{55em}}{\textbf{\modelname~{}}: <think>嗯，用户问的是行拱手礼的时候应该哪只手在外。首先，我需要回忆一下拱手礼的正确做法。拱手礼是中国传统礼仪，可能涉及到不同场合和不同手的位置。我记得可能有不同的说法，有的说左手在外，有的说右手，或者根据性别不同而不同？

首先，我应该确认传统礼仪中的正确做法。可能古代文献或者现代礼仪指南中有明确的说明。比如，古代行拱手礼时，通常是左手在外，右手在内，因为左为尊，尤其是在古代，左手被视为更尊重的一侧。不过，也有可能不同朝代有不同的习惯，或者不同地区有不同的做法。

另外，可能要考虑行礼的场合，比如是祭祀、拜见长辈还是日常问候。不同场合可能有不同的规范。比如，祭祀时可能更严格，而日常问候可能比较随意。但用户的问题没有提到具体场合，所以可能需要给出一般性的回答。

...

另外，拱手礼时，手心应向下还是向内？可能手心向内，表示诚意，或者手心向下表示谦逊。需要确认这些细节，但用户的问题主要是哪只手在外，所以重点在左右手的摆放。

总结：传统拱手礼中，左手在外，右手在内，适用于男性；女性可能右手在外，但现代可能统一为左手在外。需要指出这一点，并说明现代常见的做法，可能更倾向于左手在外，同时提及传统中的性别差异，但现代可能简化。</think>

行拱手礼时，传统上**左手在外，右手在内**，这一规范源于中国古代的礼仪文化。以下是具体说明：

1. **传统规范**

   - **左手为尊**：古代以“左”为尊贵方位（如“左丞相”高于“右丞相”），因此行拱手礼时，左手在外、右手在内，象征对对方的尊重。

   - **手心方向**：双手合抱时，手心应相对或向下（而非向上），以示谦逊和诚意。

2. **性别差异（传统）**
   
   - **男性**：通常左手在外，右手在内。
   
   - **女性**：传统上部分场合可能右手在外，左手在内（如某些祭祀礼仪），但这一差异在现代已逐渐淡化。

3. **现代简化**
   
   - 当代礼仪中，拱手礼更注重形式统一，**普遍采用左手在外**的规范，性别区分较少。日常使用或非正式场合，也可根据个人习惯调整，但需保持双手对称、自然。

4. **特殊场合**
   
   - **祭祀或正式礼仪**：需严格遵循传统，左手在外，右手在内。
  
   - **国际交往**：可简化为双手合抱、掌心相对，避免过度拘泥细节。

**总结**：\textcolor{red}{现代行拱手礼时，建议以**左手在外、右手在内**为标准，既符合传统又便于操作。若遇特殊礼仪场合，可提前了解具体要求。}
    } \\
    \bottomrule
    \end{tabular}%
    }
  \label{tab:aigc-case-1}%
\end{table*}%
\end{CJK*}

\end{CJK}
\end{document}